\documentclass[final,5p,times,twocolumn]{elsarticle}

\usepackage{amssymb}
\usepackage{amsmath,amsfonts}

\usepackage{algorithmic}
\usepackage{algorithm}
\usepackage{array}
\usepackage{booktabs}
\usepackage{textcomp}
\usepackage{changes}
\usepackage[table,xcdraw]{xcolor}
\usepackage{url}
\usepackage{verbatim}
\usepackage{hyperref}
\usepackage{multirow}
\usepackage{graphicx}
\usepackage{tikz}
\definecolor{saOneColor}{RGB}{26, 111, 168}
\definecolor{saTwoColor}{RGB}{42, 122, 59}
\definecolor{saThreeColor}{RGB}{133, 100, 4}
\newcommand{\saBadge}[2]{\tikz{\node[circle,fill=#1,text=white,font=\small\bfseries,inner sep=1.5pt,minimum size=14pt]{#2};}}
\usepackage{subcaption}
\usepackage{wasysym}

\biboptions{sort&compress}

\journal{Robotics and Computer-Integrated Manufacturing}

\begin{document}

\begin{frontmatter}



\title{LABIT: Long-Horizon Robotic Assembly Benchmark for Industrial Tasks}


\author{Constantin Schempp\corref{cor1}}
\ead{constantin.schempp@h-ka.de}
\author{Christian Friedrich}
\author{Luca Kofler}
\author{Laura Suhm}
\cortext[cor1]{Corresponding author}

\affiliation{organization={Institute for Robotics and Intelligent Production Systems, University of Applied Sciences Karlsruhe (HKA)},
            addressline={Moltkestraße 30},
            city={Karlsruhe},
            postcode={76133},
            country={Germany}}

\begin{abstract}
Robot learning has undergone significant advancements in the domain of compliant control and robotic assembly.
The development of novel methodologies has transcended the capacity to address discrete tasks, such as grasping or inserting, and has evolved to encompass the comprehensive construction of entire assemblies.
Conventional benchmarks, which are predominantly focused on single tasks, are no longer adequate to address the complexity and interconnected nature of modern manufacturing processes.
Consequently, the necessity for benchmarks to encompass a broader spectrum of assembly and subassembly steps is becoming apparent.
In light of these developments, we propose LABIT, an industrial robotic assembly benchmark for long-horizon manipulation.
This benchmark encompasses DIN-norm assembly tasks, including screwing, placing, inserting, and snapping, under both high and low tolerance conditions.
The successive nature of these assembly steps implies that prior errors can influence subsequent manipulation.
Consequently, this benchmark is particularly relevant for assessing the robustness of intelligent robotic systems.
The proposed benchmarking protocols and novel performance metric, based on a taxonomy of assembly complexity in conjunction with the comprehensive list of parts and their corresponding CAD models, serve as essential tools for research and development in the field.
To provide a baseline, we evaluate the benchmark using traditional force-position control and report reliability and score metrics per primitive, subtask, assembly category, and subassembly.
Accompanying data is available at
{\color{blue}{\href{https://irp-hka.github.io/robotic-assembly-benchmark/}{https://irp-hka.github.io/robotic-assembly-benchmark/}}}.
\end{abstract}



\begin{keyword}
benchmark \sep assembly \sep long-horizon \sep simulation \sep evaluation \sep robotics \sep industry
\end{keyword}

\end{frontmatter}



\section{Introduction}
The manufacturing industry is still a key sector in high-wage countries today.
It makes up 15.2\% of the world's gross value \cite{GDP}. Assembly is an important task within a manufacturing process and despite the progress in research, it remains one of the biggest cost drivers. Furthermore, a decline in the labor force due to a shrinking population can be observed in western countries.
This demands an increase in automation, which can be achieved using intelligent robotic systems.
To solve this issue of robot-guided industrial assembly, a full stack of robotics research is required.
This includes topics such as planning, compliant control, perception, grasping and failure recovery.
Because this is a challenging problem, it has garnered significant attention from the research community.
\begin{figure}
    \centering
    \includegraphics[width=0.48\textwidth]{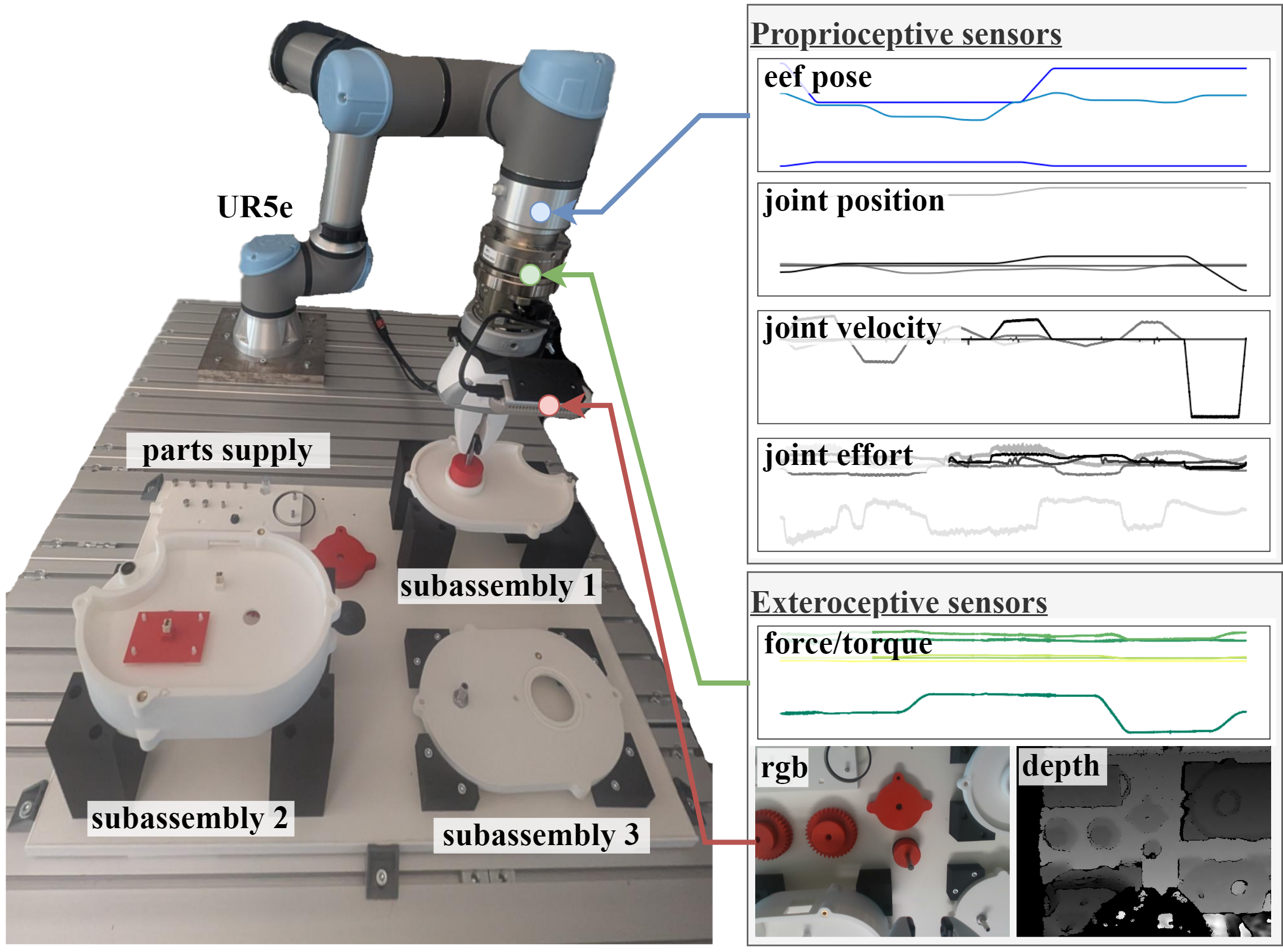}
    \caption{Experimental setup for LABIT. A robot arm equipped with a parallel-jaw gripper, a wrist-mounted RGB-D camera, and a force-torque sensor executes the assembly tasks on a dedicated worktable. The benchmark housing is secured in a fixture; individual parts are provisioned via a part feeder, corresponding to planning complexity level~1.}
    \label{fig:experimental_setup}
\end{figure}

\begin{figure*}[!t]
    \centering
    \begin{subfigure}[b]{0.5\textwidth}
        \centering
        \captionsetup{justification=centering}
        \begin{minipage}[t]{0.32\linewidth}
            \centering
            \includegraphics[width=\linewidth]{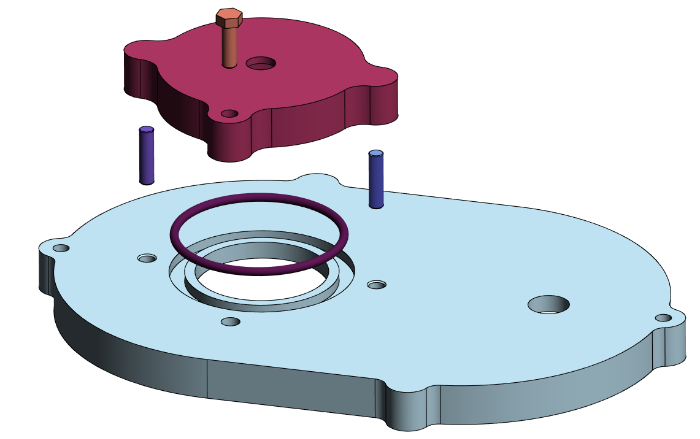}
        \end{minipage}%
        \begin{minipage}[t]{0.32\linewidth}
            \centering
            \includegraphics[width=\linewidth]{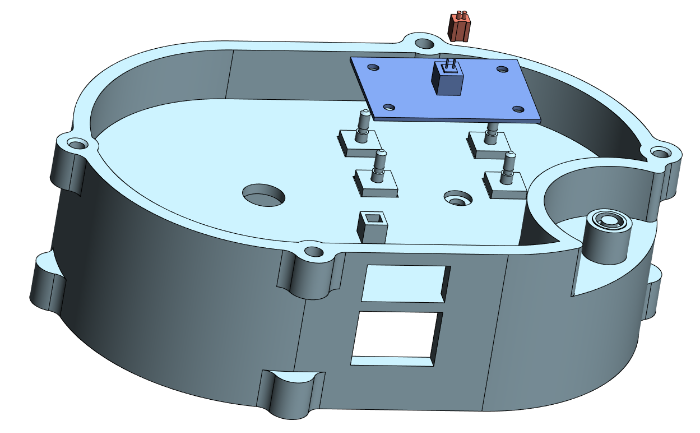}
        \end{minipage}%
        \begin{minipage}[t]{0.32\linewidth}
            \centering
            \includegraphics[width=\linewidth]{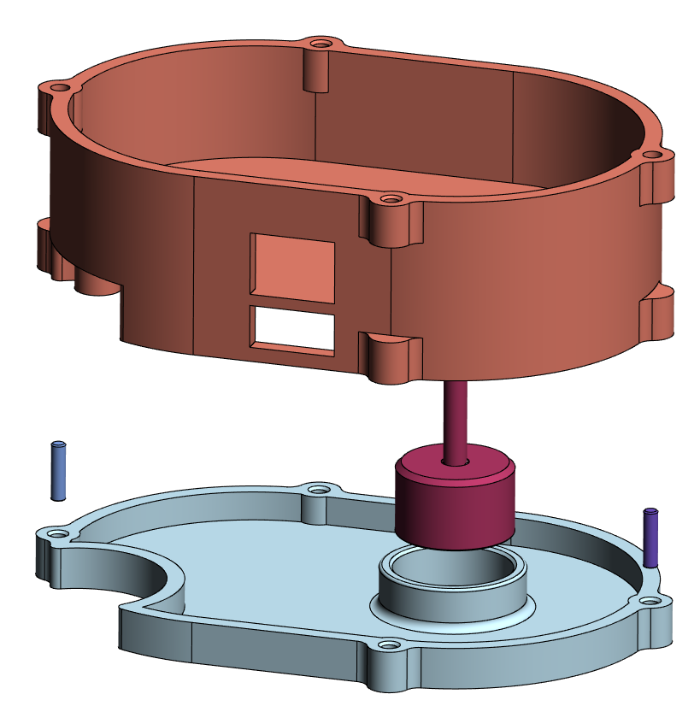}
        \end{minipage}%
        \vspace{0.7cm}

        \begin{minipage}[t]{0.32\linewidth}
            \centering
            \includegraphics[width=\linewidth]{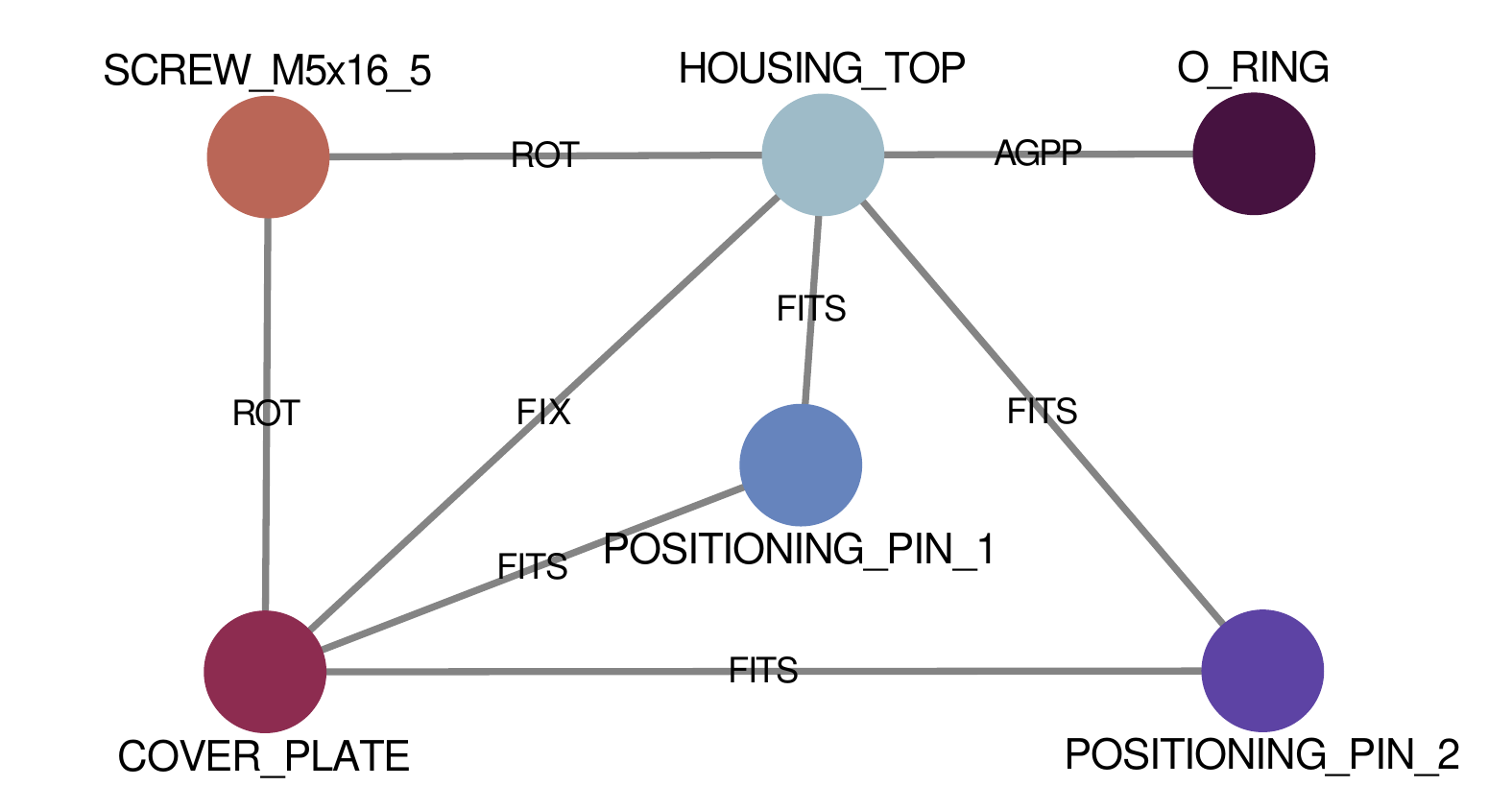}\\[3pt]
            \saBadge{saOneColor}{1}
        \end{minipage}
        \begin{minipage}[t]{0.32\linewidth}
            \centering
            \includegraphics[width=\linewidth]{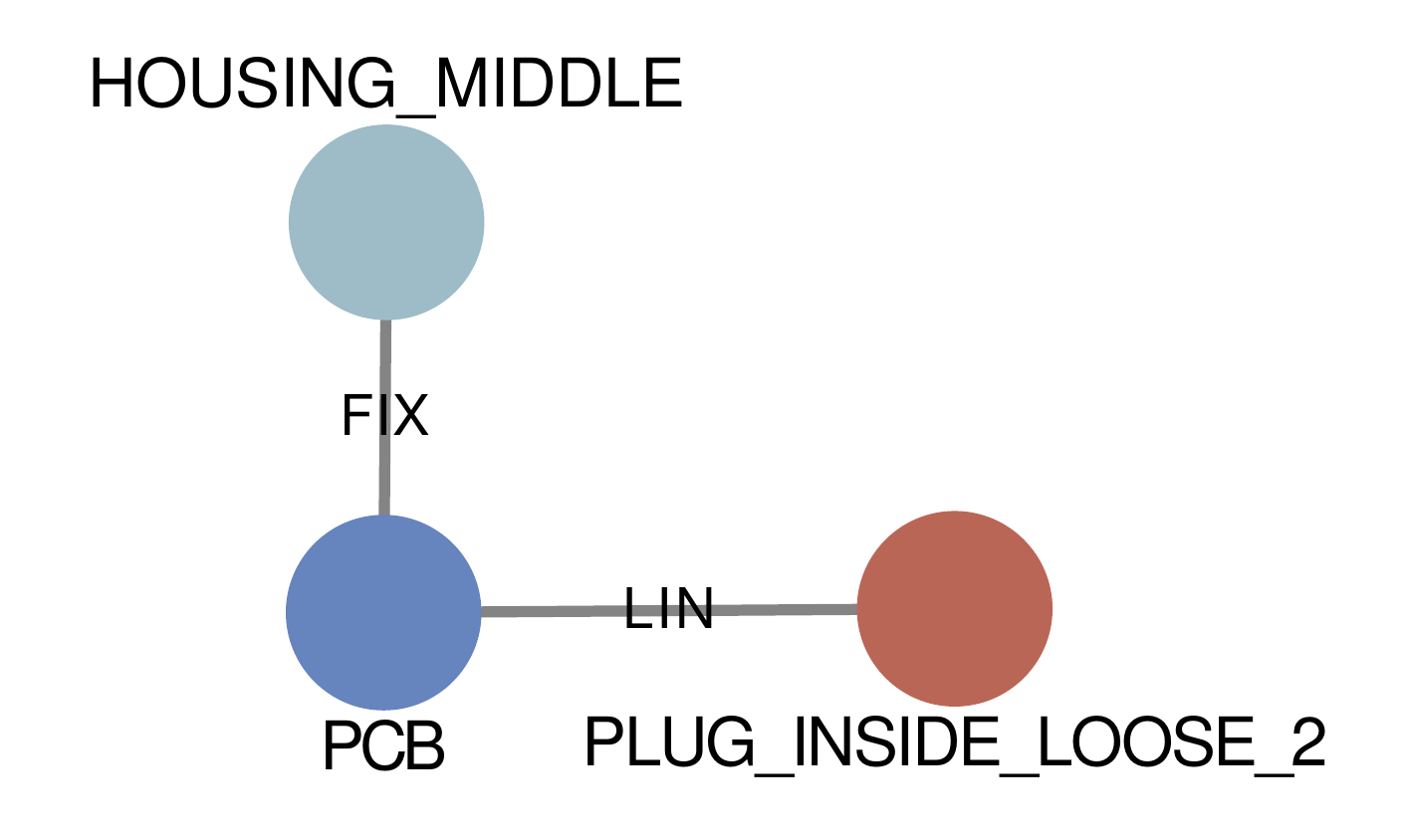}\\[3pt]
            \saBadge{saTwoColor}{2}
        \end{minipage}
        \begin{minipage}[t]{0.32\linewidth}
            \centering
            \includegraphics[width=\linewidth]{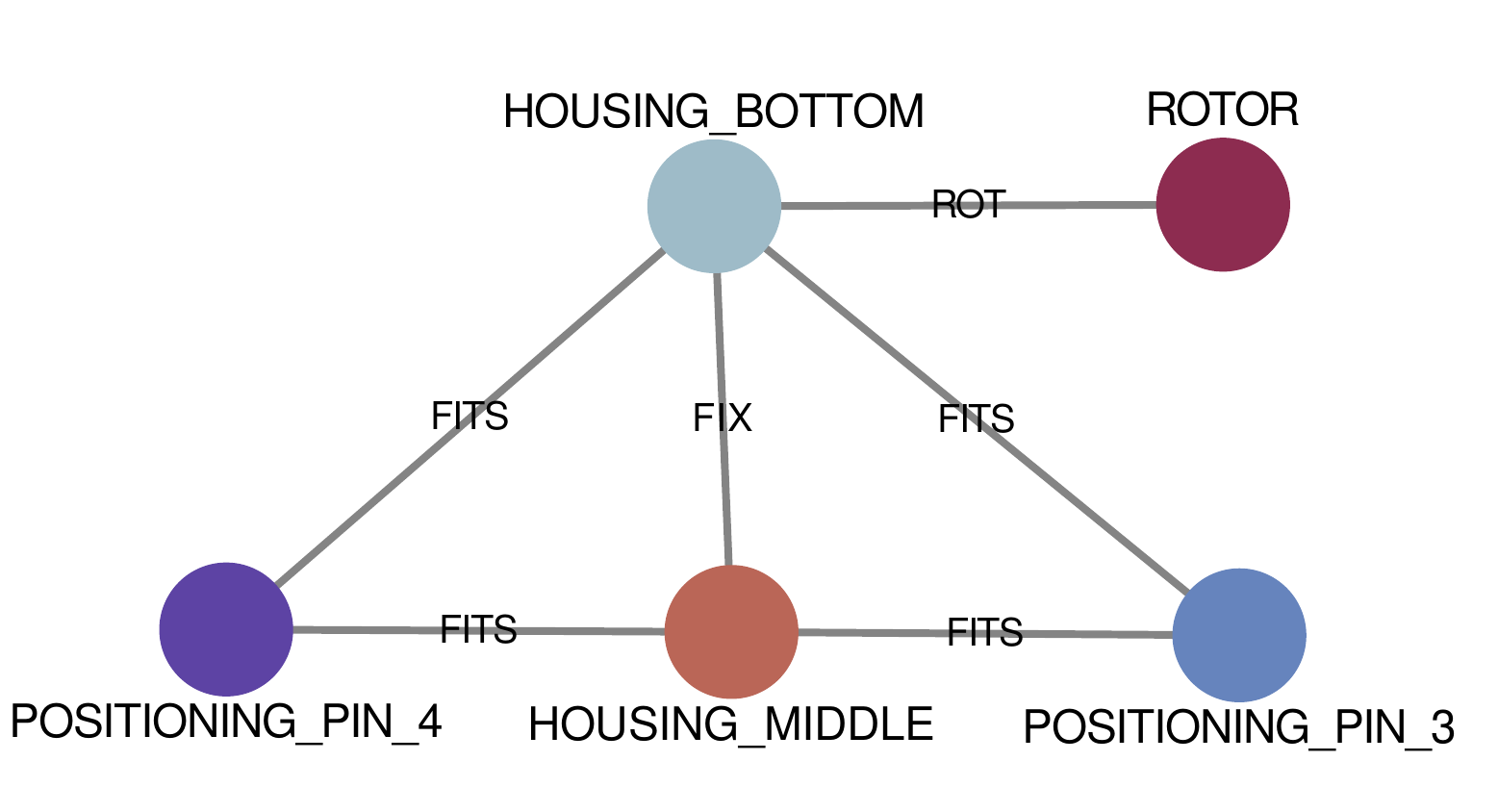}\\[3pt]
            \saBadge{saThreeColor}{3}
        \end{minipage}
        \caption{(1) cover plate, (2) PCB module, (3) middle housing.}
    \end{subfigure}%
    \begin{subfigure}[b]{0.5\textwidth}
        \centering
        \captionsetup{justification=centering}
        \includegraphics[width=\linewidth]{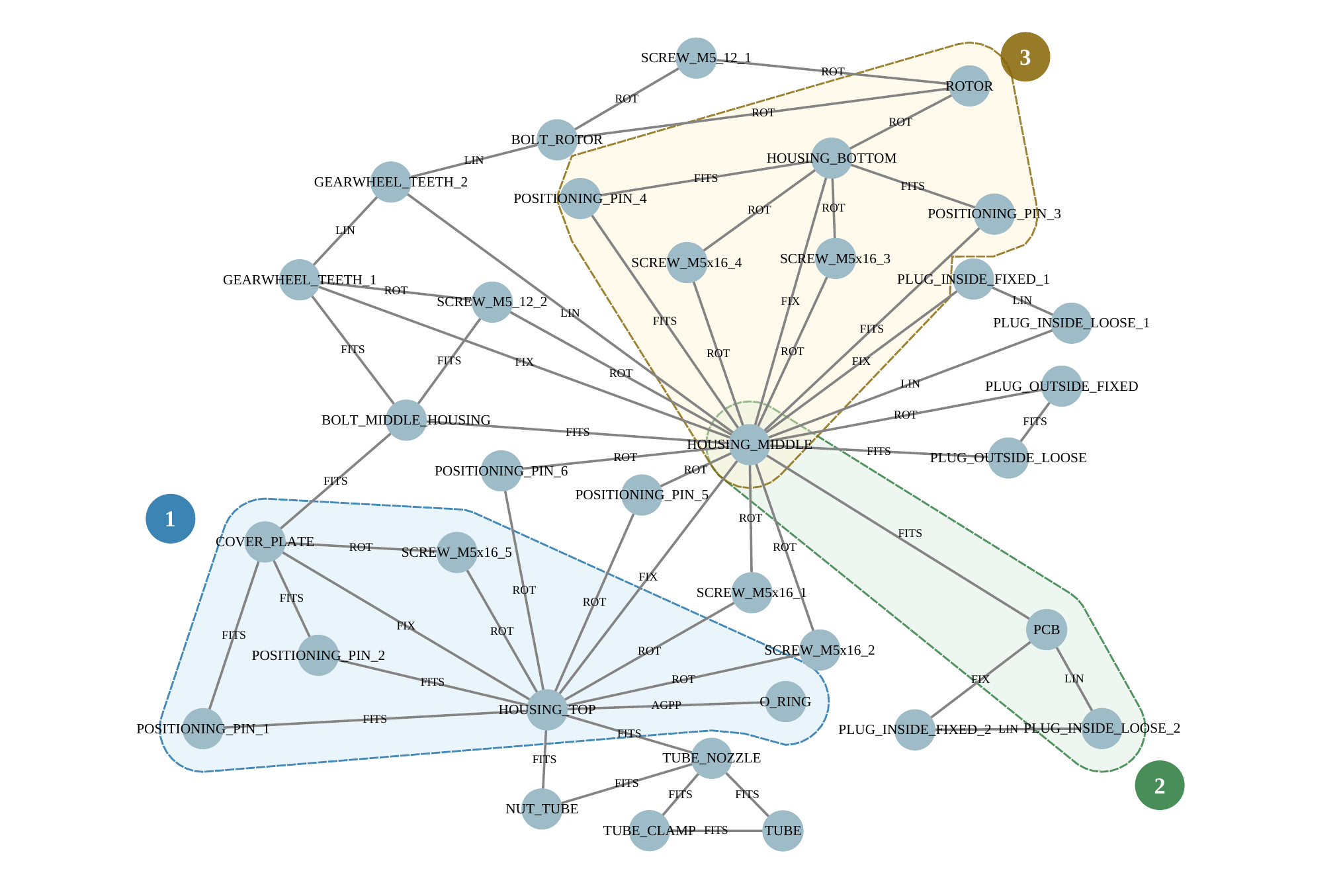}
        \caption{Top-level assembly graph of LABIT.}
    \end{subfigure}
    \caption{Hierarchical structure of the LABIT benchmark. 
    (a) shows explosion views (top row) and corresponding primitive-level assembly graphs (bottom row) for three exemplary assemblies: cover plate (left), PCB module (center), and middle housing (right). 
        Nodes represent individual assembly parts, where node colors match with part colors in the explosion views; edges display the spatial relationship and type of connections. The SCREW\_M5x16 nodes of the housing middle assembly are not depicted here, because they are part of a later assembly stage.
    (b) shows the top-level graph connecting all assembly parts from LABIT, where the highlighted areas correspond to the assemblies shown in (a).
    }\label{fig:overview_benchmark}
\end{figure*}

A broad range of methods has been proposed for robot-guided assembly.
On the reinforcement learning (RL) side~\cite{ElgueaAguinaco_2023}, representative approaches include force-guided RL for high-precision insertion~\cite{Luo_Li_2021}, deep RL with virtual guiding forces for peg-in-hole tasks~\cite{Zang_Wang_Pan_Hou_Ding_Zhao_2024}, and sim-to-real transfer via domain adaptation combined with compliant control~\cite{Shi_Yuan_Tsitos_Cong_Hadjar_Chen_Zhang_2023}.
Imitation learning is also applied, covering hybrid trajectory-force imitation from human demonstrations~\cite{Wang_Beltran-Hernandez_Wan_Harada_2021}, data-efficient behavior cloning for contact-rich assembly~\cite{Ankile_Simeonov_Shenfeld_Agrawal_2024}, residual imitation learning in semi-structured environments~\cite{Wang_Su_Sun_Chen_Xie_2024}, and the combination of imitation and reinforcement learning for high-precision peg-in-hole assembly~\cite{Ding_2026}.
Beyond purely learned controllers, hierarchical hybrid methods couple learning with classical motion planning to address longer horizons and contact-rich sequences~\cite{Sun_Curtis_You_Xu_Koehle_Guibas_Chitta_Schwager_Li_2024, 9784890}.

Despite this progress, these methods are difficult to reproduce and rarely evaluated against each other in equal task settings, making it hard to determine which approach is best suited for a given application.
Benchmarks have proven effective at resolving this comparability problem in related fields, having significantly accelerated progress in object detection~\cite{Deng_Dong_Socher_Li_KaiLi_LiFei-Fei_2009}, grasping~\cite{Fang_2020_CVPR}, and reinforcement learning~\cite{pmlr-v100-yu20a,james2019rlbench}.

In the field of robot-guided assembly, however, the available benchmarks are limited to specific subtasks such as insertion or screwing, and do not capture the complexity of assembling a complete product.
For real-world applications, evaluating approaches across multi-step sequences with hierarchical dependencies is far more relevant.
Consequently, there is a need for a realistic, long-horizon industrial assembly benchmark that encompasses subassemblies and their interdependencies.
\begin{table}[t]
\centering
\caption{Comparison of robotic assembly benchmarks.
\emph{Ops.}: number of distinct operation types.
\emph{Seq.}: long-horizon sequential assembly with hierarchical step dependencies (failure in step $i$ directly affects step $i{+}1$).
\emph{Ind.}: industrial off-the-shelf parts used.
\emph{Force}: force-sensitive/compliant control required.
\emph{Reor.}: explicit subassembly reorientation required.
\emph{Sim.}: simulation environment provided.
\emph{Phys.}: physical hardware provided.
\CIRCLE~yes \quad \LEFTcircle~partial \quad \Circle~no.}
\label{tab:benchmark_comparison}
\setlength{\tabcolsep}{4pt}
\resizebox{\linewidth}{!}{%
\begin{tabular}{@{}lccccccc@{}}
\toprule
Benchmark & Ops. & Seq. & Ind. & Force & Reor. & Sim. & Phys. \\ \midrule
LABIT (ours) & 7 & \CIRCLE & \LEFTcircle & \CIRCLE & \CIRCLE & \CIRCLE & \CIRCLE \\
FurnitureBench \cite{Heo_Lee_Lee_Lim_2023} & 4 & \CIRCLE & \Circle & \Circle & \LEFTcircle & \CIRCLE & \CIRCLE \\
FMB \cite{Luo_Xu_Liu_Tan_Lin_Wu_Abbeel_Levine_2024} & 3 & \LEFTcircle & \Circle & \LEFTcircle & \LEFTcircle & \CIRCLE & \CIRCLE \\
RAMP \cite{Collins_Robson_Yamada_Sridharan_Janik_Posner_2024} & 3 & \CIRCLE & \LEFTcircle & \Circle & \Circle & \CIRCLE & \CIRCLE \\
REASSEMBLE \cite{Sliwowski_Jadav_Stanovcic_Orbik_Heidersberger_Lee_2025} & 4 & \Circle & \CIRCLE & \CIRCLE & \Circle & \Circle & \CIRCLE \\
AutoMate \cite{Tang_Akinola_Xu_Wen_Handa_VanWyk_Fox_Sukhatme_Ramos_Narang_2024} & 2 & \Circle & \Circle & \Circle & \Circle & \CIRCLE & \CIRCLE \\
RoCo \cite{Liu_etal_RoCo_2026} & 3 & \LEFTcircle & \Circle & \LEFTcircle & \Circle & \CIRCLE & \CIRCLE \\
Fabrica \cite{Tian_Jacob_Huang_Zhao_Gu_Ma_Zhang_Javid_Romero_Chitta_Sueda_Li_Matusik_2025} & 2 & \CIRCLE & \Circle & \LEFTcircle & \Circle & \CIRCLE & \CIRCLE \\
\bottomrule
\end{tabular}}
\vspace{-5mm}
\end{table}

\section{Related Work}
Benchmarks play a crucial role in advancing research by enabling the systematic evaluation of progress and facilitating direct comparisons between different approaches. Over the years, various robotic benchmarks have been proposed, covering a wide range of tasks. However, relatively few of these focus specifically on assembly, despite its significance in real-world manufacturing and automation. Recently, several benchmarks have emerged that address this gap, providing structured evaluations for robotic assembly tasks.

Among the existing assembly-focused benchmarks, many are designed for short-horizon tasks that involve single-step executions. Notable examples include the NIST Assembly Task Board benchmark \cite{Kimble_VanWyk_Falco_Messina_Sun_Shibata_Uemura_Yokokohji_2020}, which evaluates basic assembly operations. Additionally, some benchmarks are exclusively simulation-based, such as BulletArm \cite{Wang_Kohler_Zhu_Jia_Platt_2023} and the IKEA Furniture Assembly environment \cite{Lee_Hu_Lim_2021}, which test robotic performance in constrained virtual settings.

For long-horizon robotic assembly, where multiple sequential steps must be executed, more comprehensive benchmarks have recently been introduced. RAMP \cite{Collins_Robson_Yamada_Sridharan_Janik_Posner_2024}, FurnitureBench \cite{Heo_Lee_Lee_Lim_2023}, and FMB \cite{Luo_Xu_Liu_Tan_Lin_Wu_Abbeel_Levine_2024} provide structured environments to evaluate robotic agents across extended assembly sequences. Furthermore, real-world competitions such as the World Robot Summit Assembly Challenge \cite{Yokokohji_Kawai_et_al_2022} have encouraged the development of advanced assembly solutions, while works such as \cite{Chatzilygeroudis_Fichera_Lauzana_Bu_Yao_Khadivar_Billard_2020} explore the complexities of coordinated multi-arm assembly.
Beyond assembly-specific settings, broader research on multi-robot task allocation has addressed diverse application scenarios with varying environmental constraints and system configurations \cite{Bai_Jiang_Li_Ullah_Dabel_Bashir_Wu_Ge_2025, Bai_Li_Li_Khan_Zhang_Zhang_2023, Bai_Yan_Cao_Xue_2019, MuhammadYasirNaeem_Ahmed_AwaisButt_IqbalBhatti_Qiu_Bai_2025}.
Despite these advancements, existing benchmarks differ substantially from LABIT in their scope, part fidelity, and evaluation depth.
Table~\ref{tab:benchmark_comparison} provides a structured comparison.
Key distinguishing features of LABIT are the use of off-the-shelf industrial components (rather than exclusively 3D-printed parts), 
an explicit force-sensitive assembly regime, and a paired simulation environment together with physical hardware.
The requirement of subassembly reorientation is also a unique aspect of LABIT, as it necessitates more complex manipulation strategies and planning.

We therefore propose a robotic manipulation benchmark to close this gap, providing:
\begin{itemize}
\item subassemblies with designated sequences to enable the testing of long-horizon planning methods
\item a realistic representation of an industrial task
\item open-source data and easy-to-replicate setup
\item a newly proposed taxonomy, addressing assembly complexity, which provides general metrics for the evaluation of assembly benchmarks.
\end{itemize}

\section{Benchmark}
\begin{figure*}[ht]
    \begin{subfigure}[t]{0.3\textwidth}
        \centering
        \captionsetup{justification=centering}
        \includegraphics[width=\textwidth]{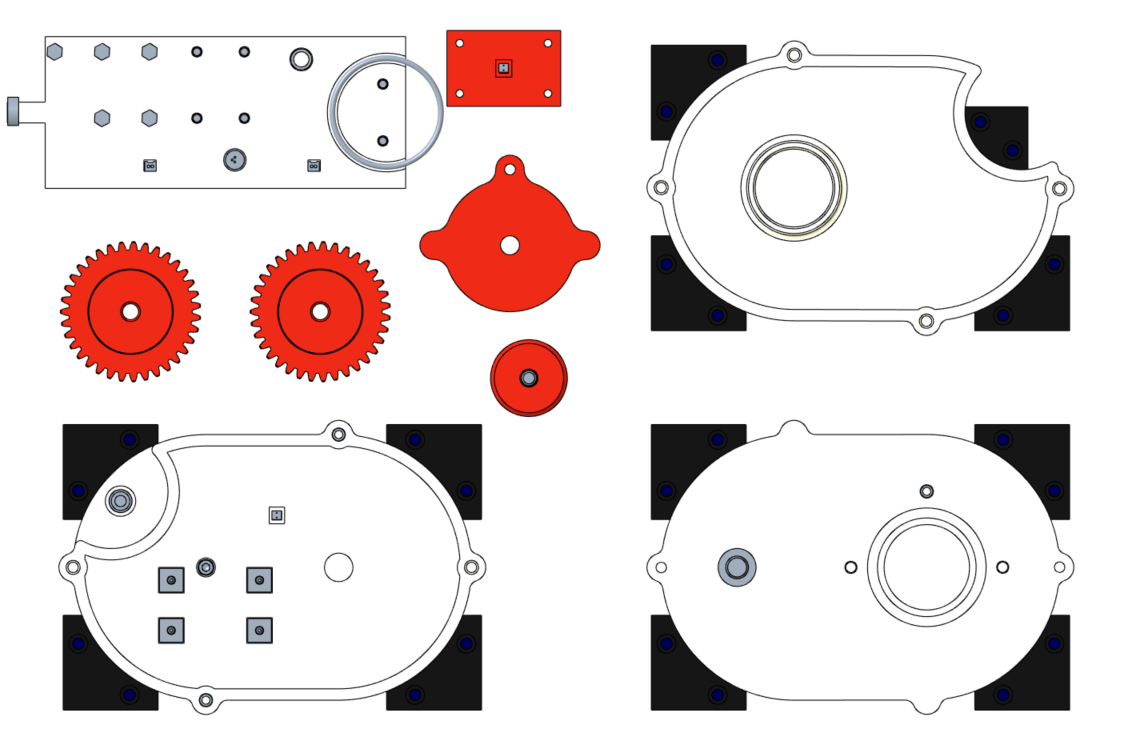}
        \caption{Complexity level 1: part feeder}
    \end{subfigure}\hfill
    \begin{subfigure}[t]{0.3\textwidth}
        \centering
        \captionsetup{justification=centering}
        \includegraphics[width=\textwidth]{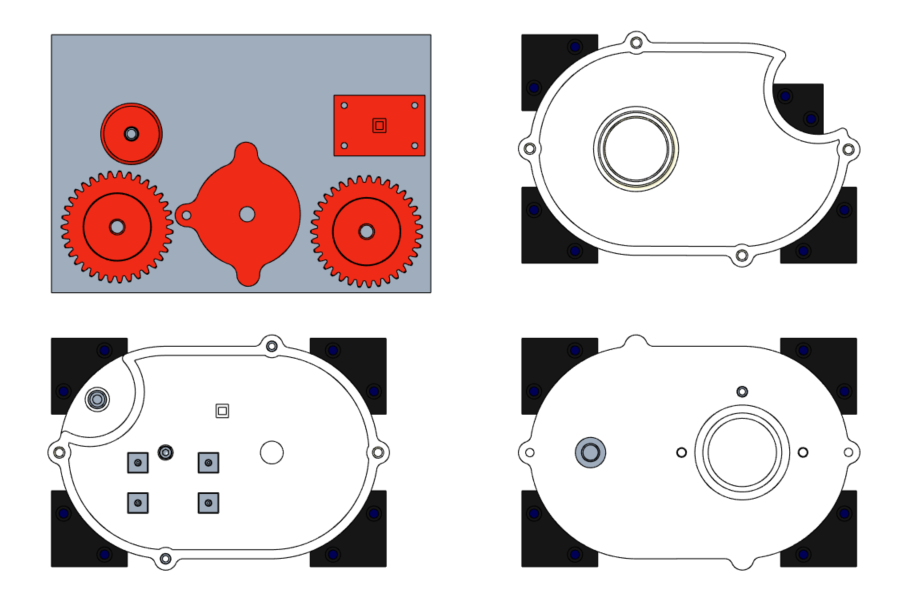}
        \caption{Complexity level 2: plate}
    \end{subfigure}\hfill
    \begin{subfigure}[t]{0.3\textwidth}
        \centering
        \captionsetup{justification=centering}
        \includegraphics[width=\textwidth]{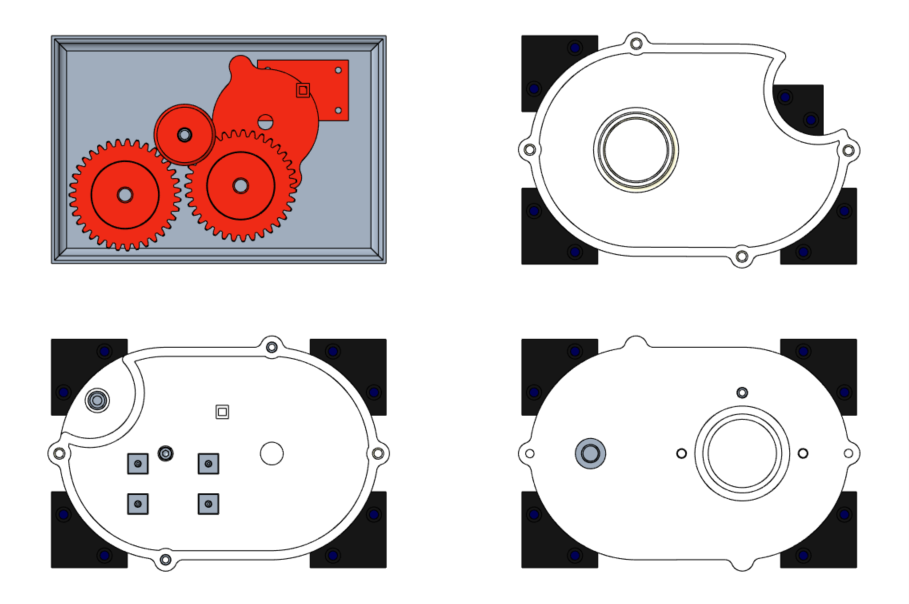}
        \caption{Complexity level 3: box}
    \end{subfigure}
\caption{Different initial benchmark setup depending on the chosen complexity level. (a) level 1, parts are provided by a part feeder, (b) level 2, parts are provided unoriented on a plate and (c) level 3, parts are provided cluttered in a box.}
\label{fig:planning_complexity}
\end{figure*}

\begin{figure*}
\vspace{2.4mm}
\centering

\begin{subfigure}[t]{0.19\textwidth}
    \captionsetup{justification=centering}
    \includegraphics[width=\textwidth]{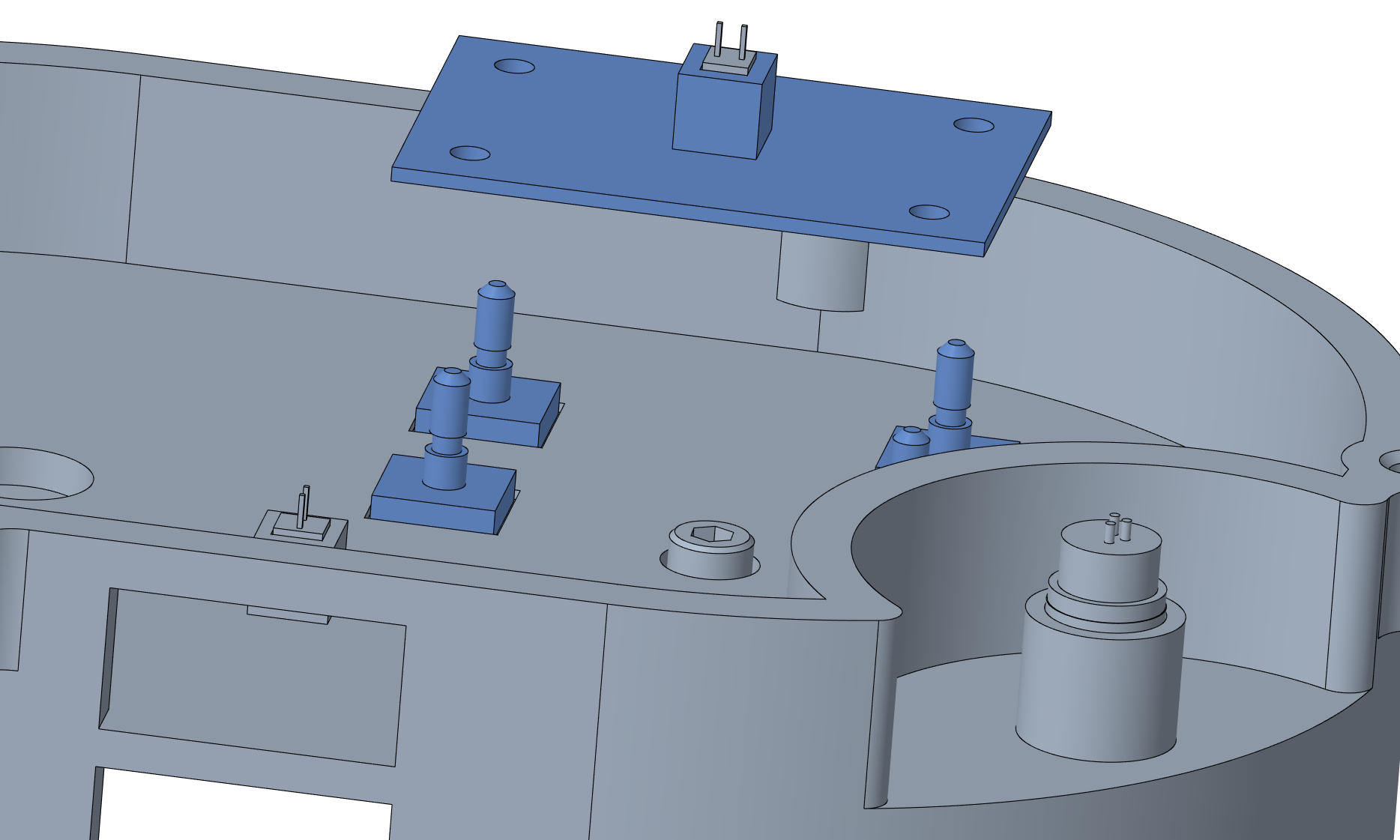}
    \caption{insert spring-loaded}
    \label{fig:benchmark_circuit_board}
\end{subfigure}
\begin{subfigure}[t]{0.19\textwidth}
    \captionsetup{justification=centering}
    \includegraphics[width=\textwidth]{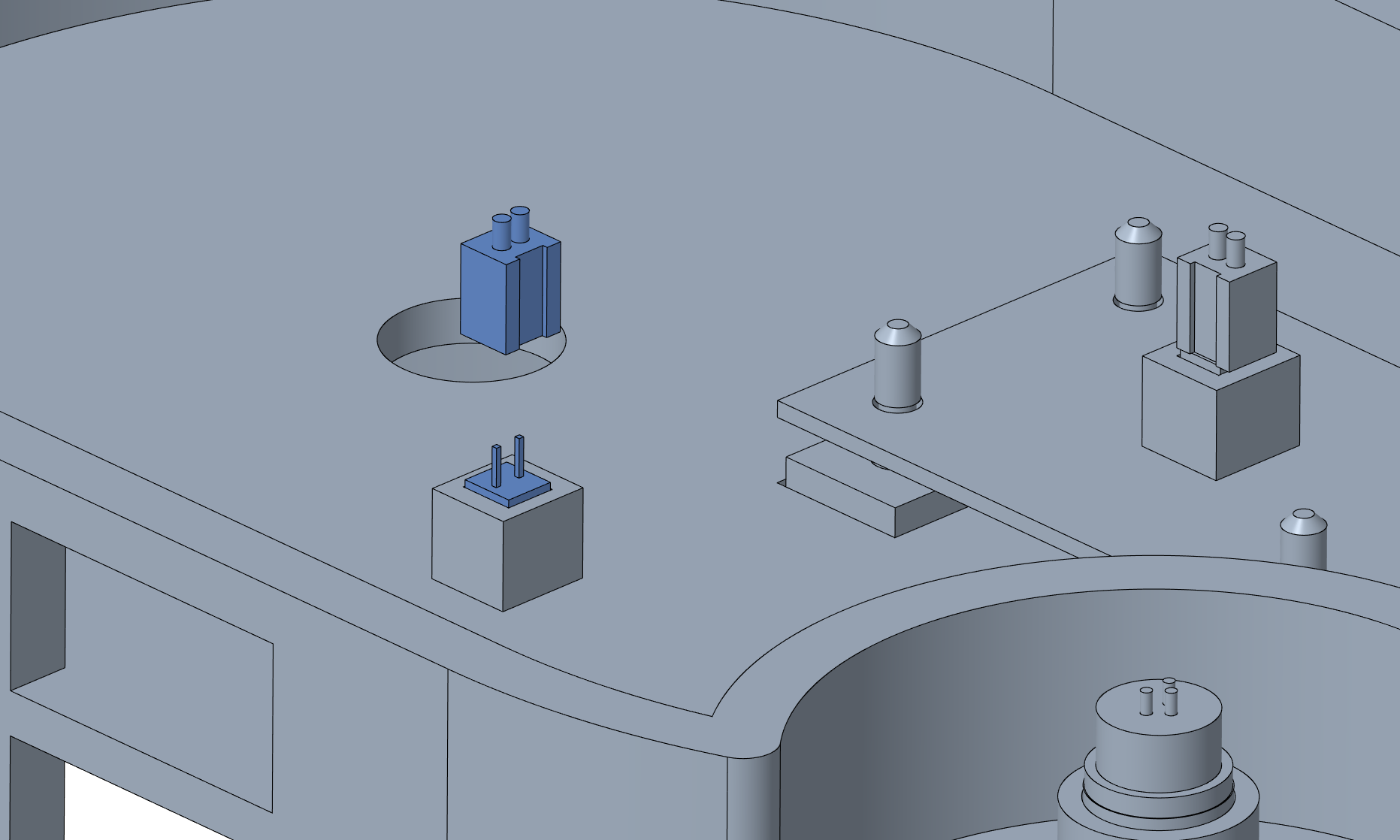}
	\caption{insert internal plug}
    \label{fig:benchmark_internal_plug}
\end{subfigure}
\begin{subfigure}[t]{0.19\textwidth}
    \captionsetup{justification=centering}
    \includegraphics[width=\textwidth]{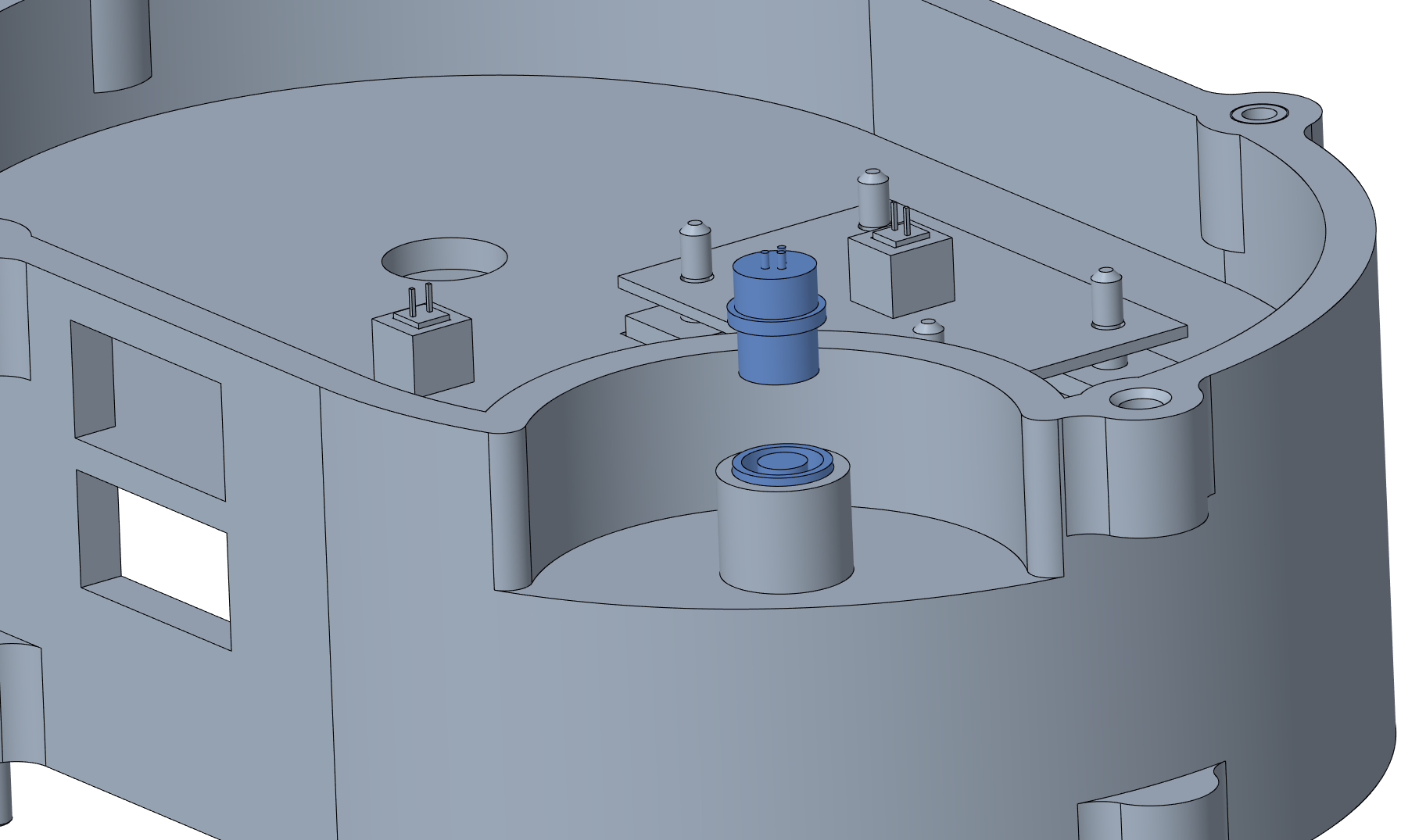}
	\caption{insert external plug}
    \label{fig:benchmark_external_plug}
\end{subfigure}
\begin{subfigure}[t]{0.19\textwidth}
    \captionsetup{justification=centering}
    \includegraphics[width=\textwidth]{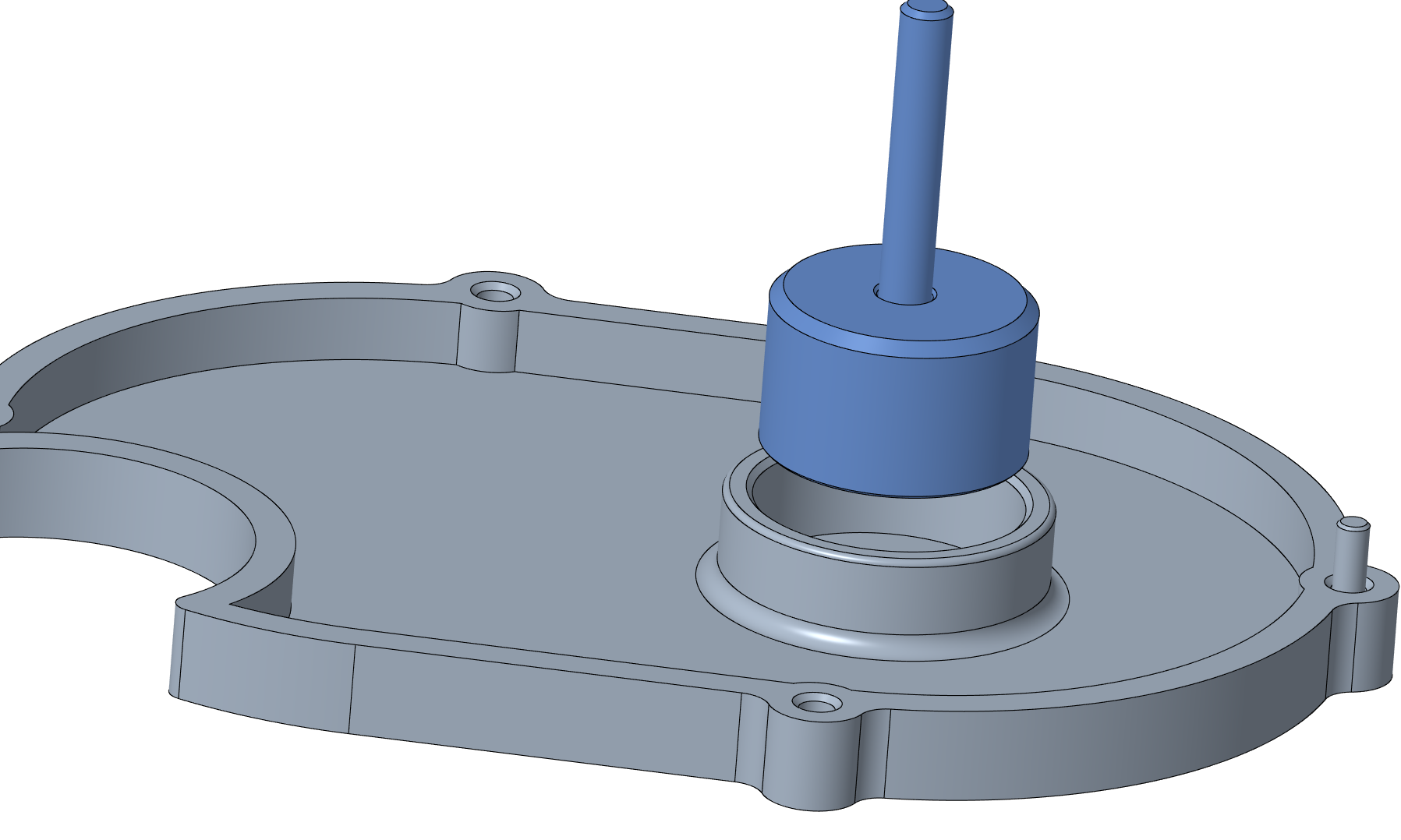}
	\caption{insert rotor}
    \label{fig:benchmark_rotor}
\end{subfigure}
\begin{subfigure}[t]{0.19\textwidth}
    \captionsetup{justification=centering}
    \includegraphics[width=\textwidth]{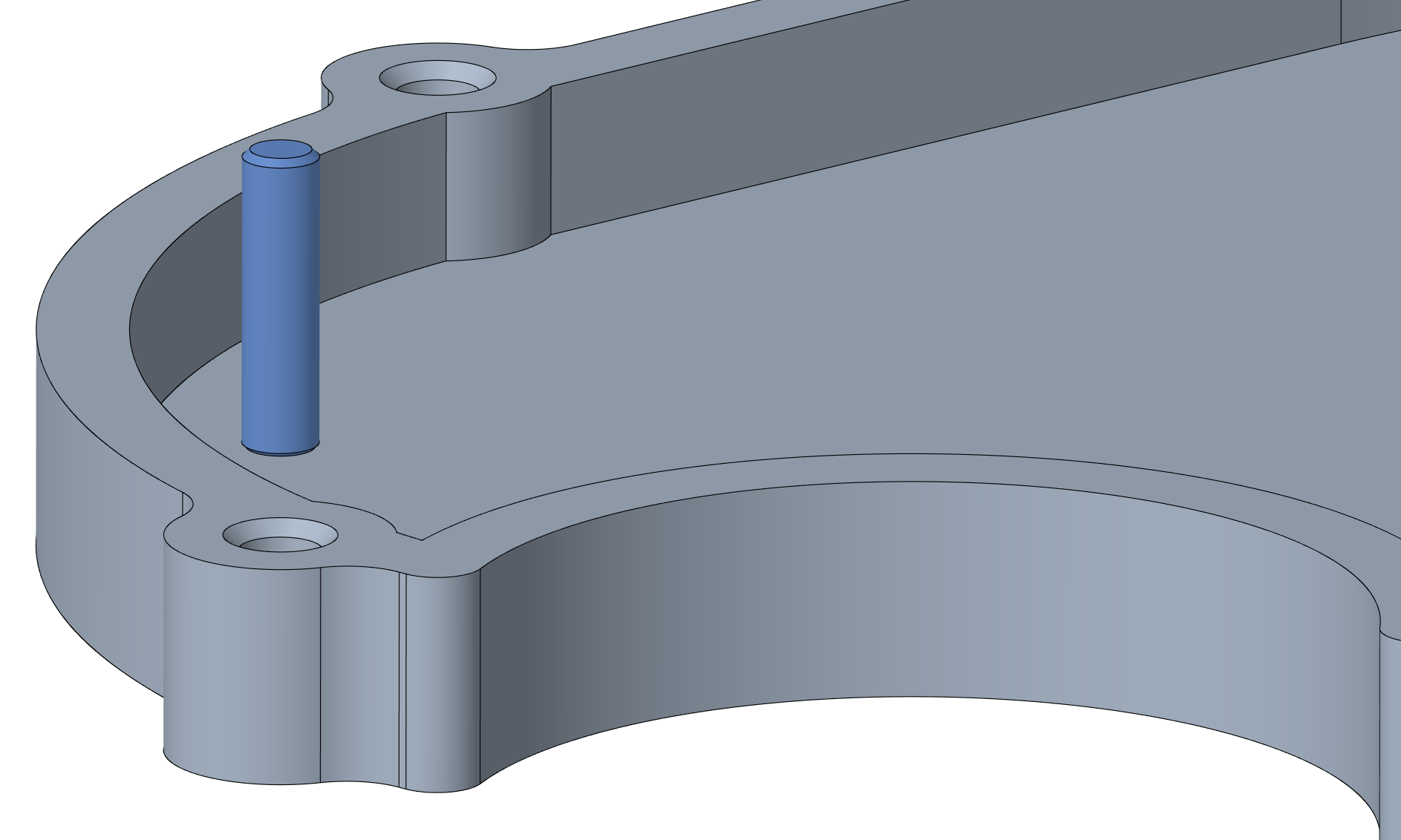}
    \caption{insert peg}
    \label{fig:benchmark_peg}
\end{subfigure}

\begin{subfigure}[t]{0.19\textwidth}
    \captionsetup{justification=centering}
    \includegraphics[width=\textwidth]{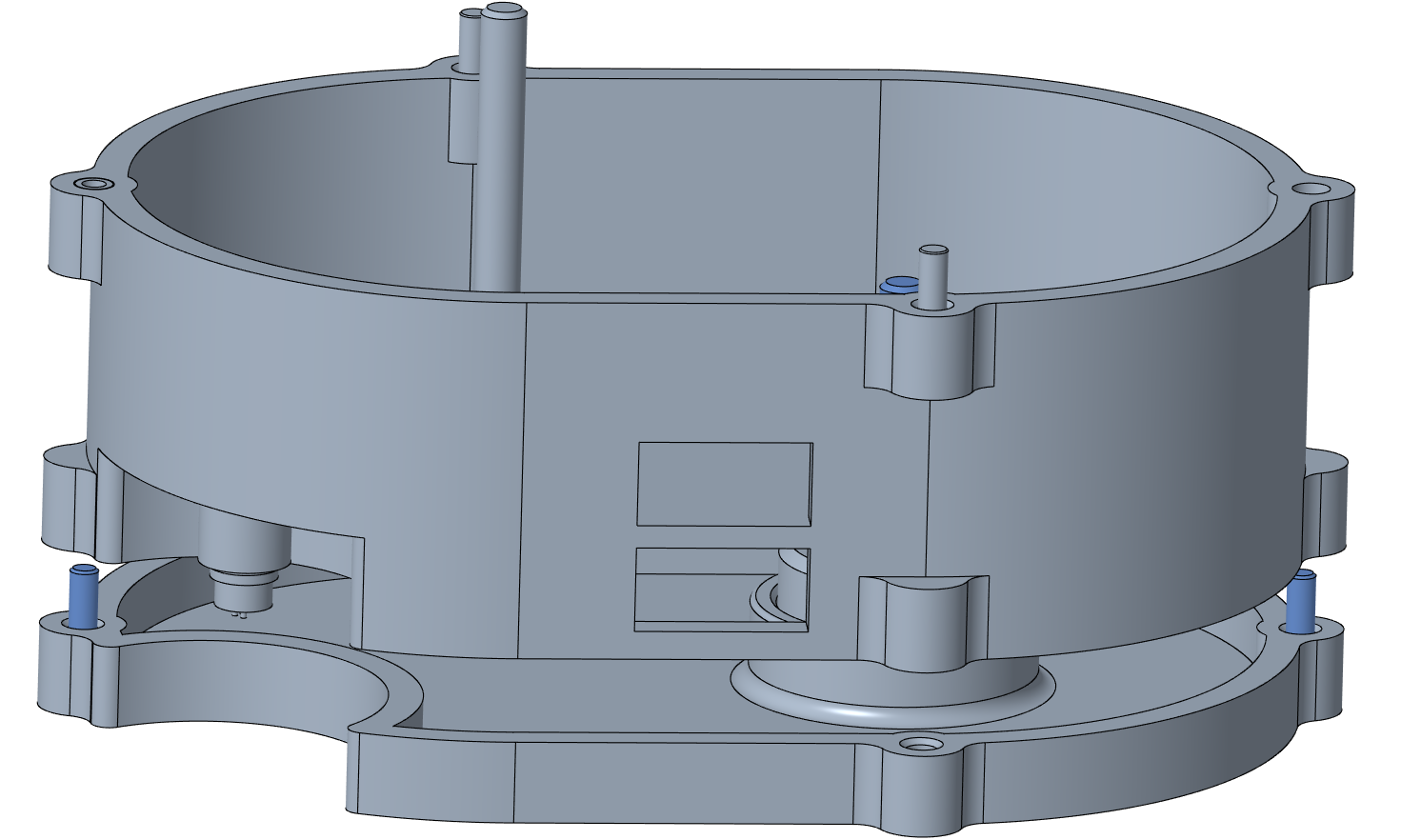}
    \caption{reorient center housing}
    \label{fig:benchmark_center_housing}
\end{subfigure}
\begin{subfigure}[t]{0.19\textwidth}
    \captionsetup{justification=centering}
    \includegraphics[width=\textwidth]{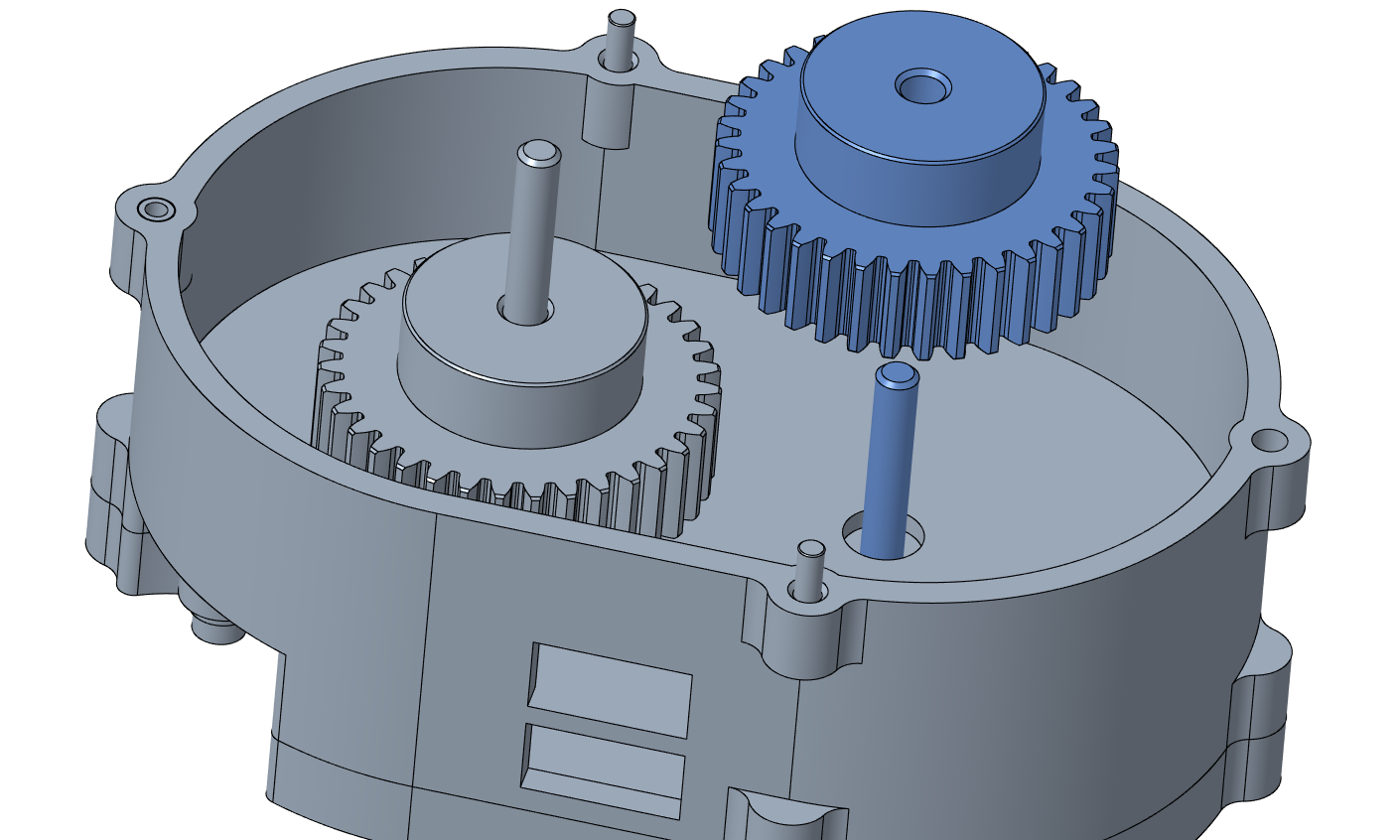}
	\caption{align gear wheel}
    \label{fig:benchmark_align_gear_wheel}
\end{subfigure}
\begin{subfigure}[t]{0.19\textwidth}
    \captionsetup{justification=centering}
    \includegraphics[width=\textwidth]{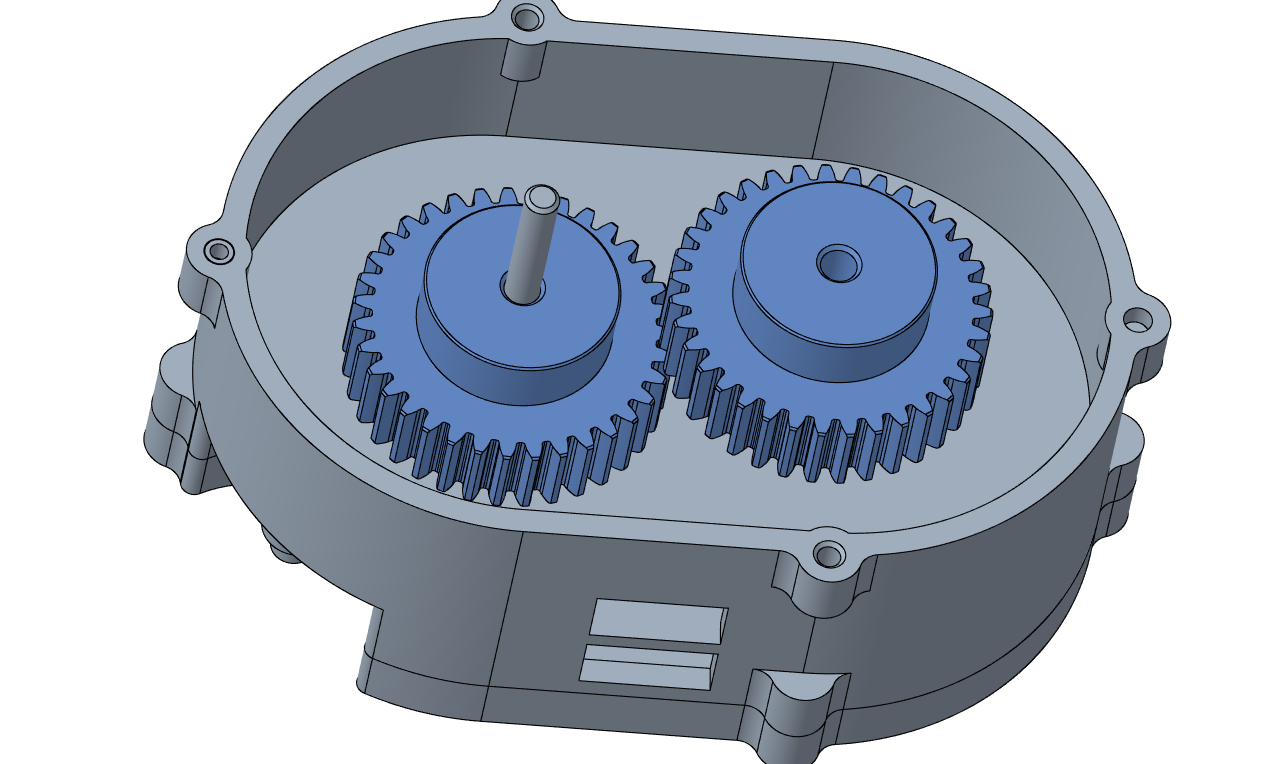}
    \caption{insert gear wheel}
    \label{fig:benchmark_insert_gear_wheel}
\end{subfigure}
\begin{subfigure}[t]{0.19\textwidth}
    \captionsetup{justification=centering}
	\includegraphics[width=\textwidth]{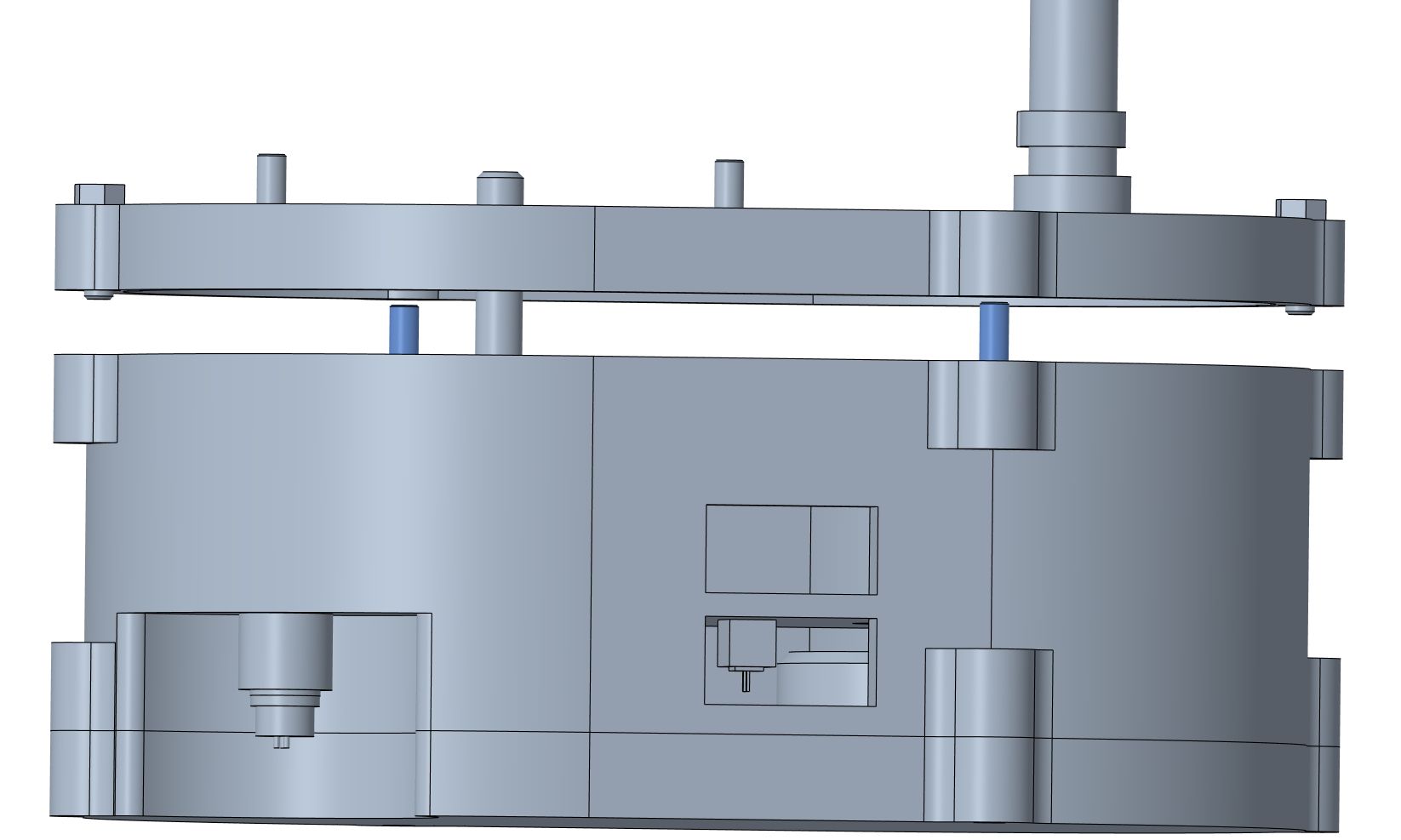}
    \caption{reorient top housing}
    \label{fig:benchmark_align_top}
\end{subfigure}
\begin{subfigure}[t]{0.19\textwidth}
    \captionsetup{justification=centering}
    \includegraphics[width=\textwidth]{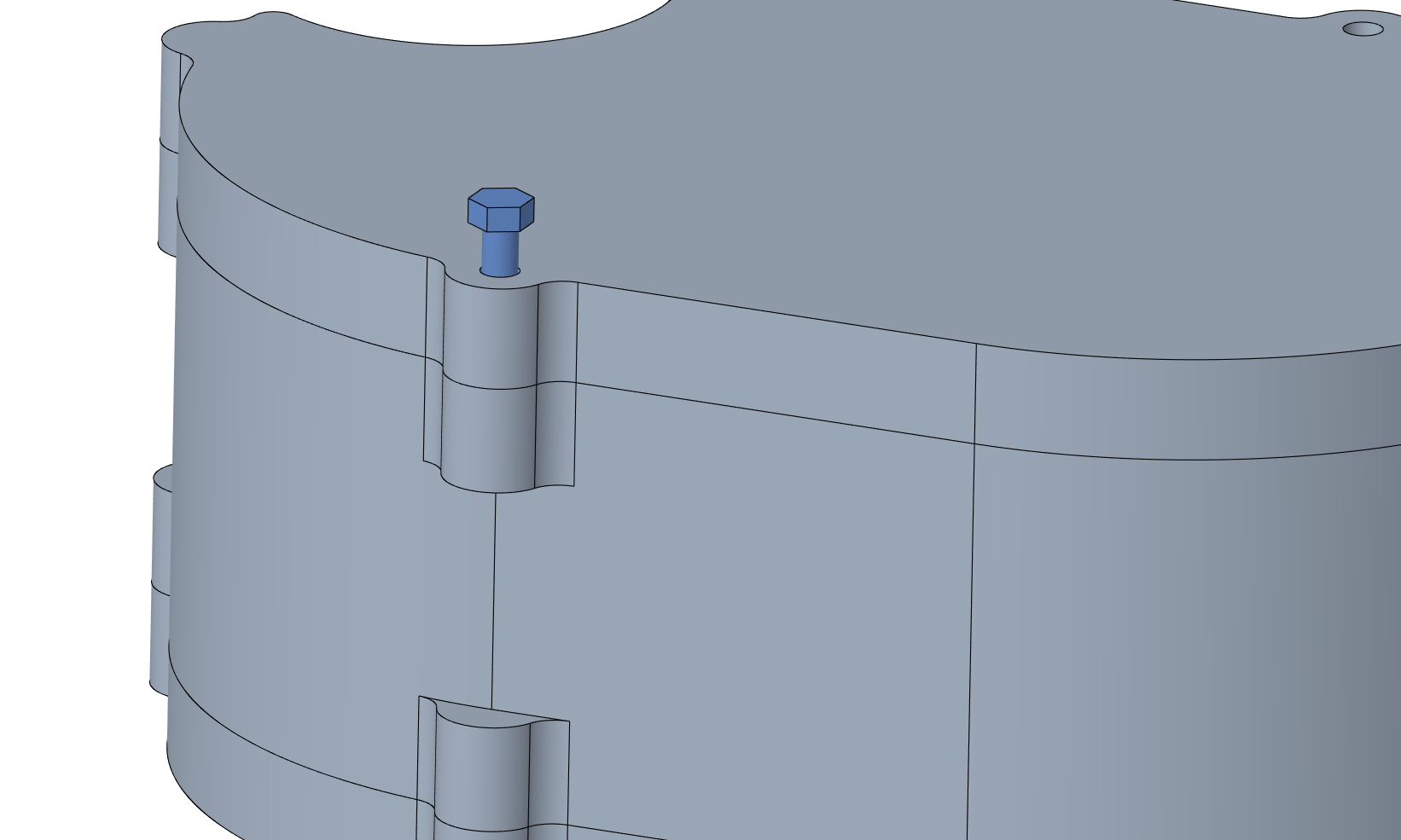}
    \caption{screwing}
    \label{fig:benchmark_screwing}
\end{subfigure}

\begin{subfigure}[t]{0.19\textwidth}
    \captionsetup{justification=centering}
    \includegraphics[width=\textwidth]{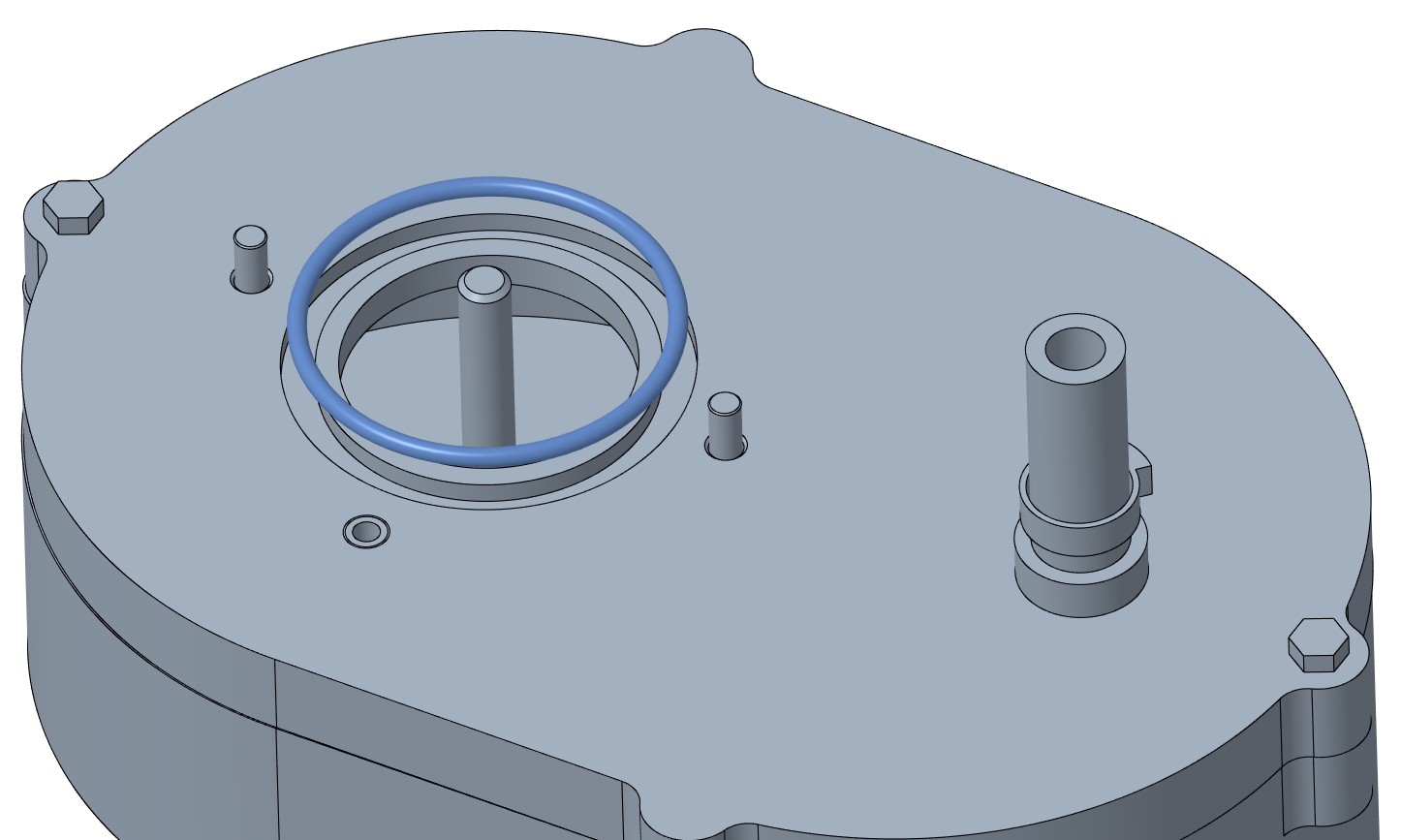}
    \caption{place o-ring}
    \label{fig:benchmark_oring}
\end{subfigure}
\begin{subfigure}[t]{0.19\textwidth}
    \captionsetup{justification=centering}
	\includegraphics[width=\textwidth]{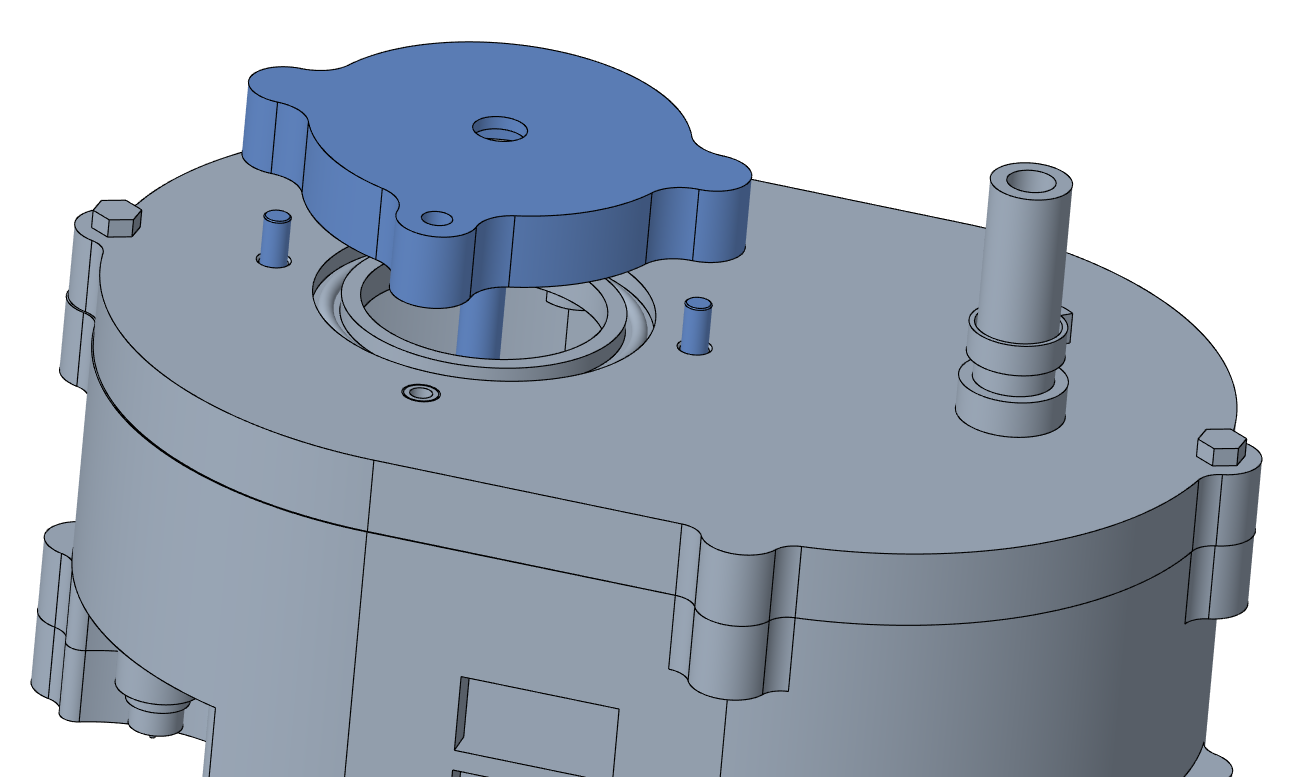}
    \caption{align cover plate}
    \label{fig:benchmark_align_coverplate}
\end{subfigure}
\begin{subfigure}[t]{0.19\textwidth}
    \captionsetup{justification=centering}
	\includegraphics[width=\textwidth]{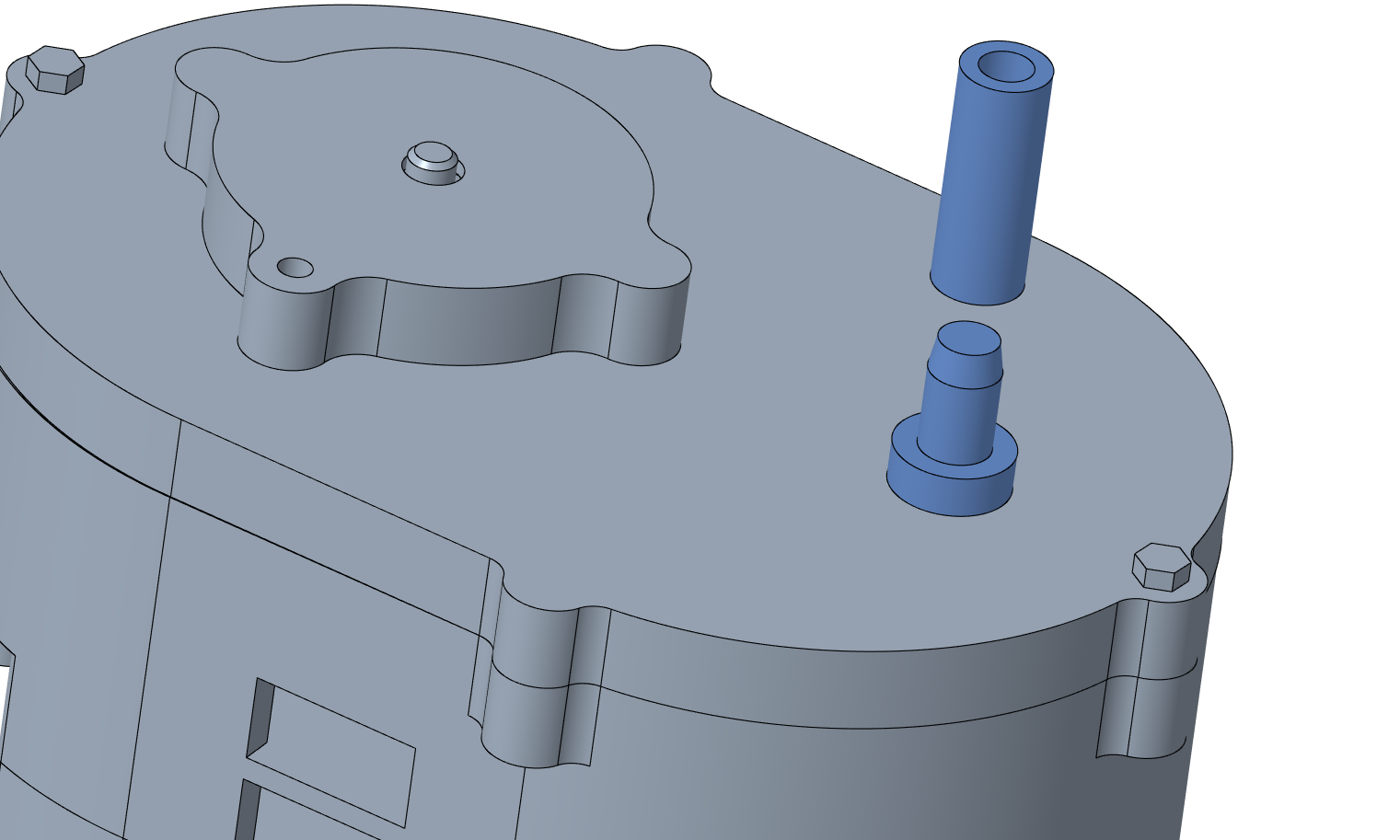}
    \caption{slide on the hose}
    \label{fig:benchmark_slide_hose}
\end{subfigure}
\begin{subfigure}[t]{0.19\textwidth}
    \captionsetup{justification=centering}
	\includegraphics[width=\textwidth]{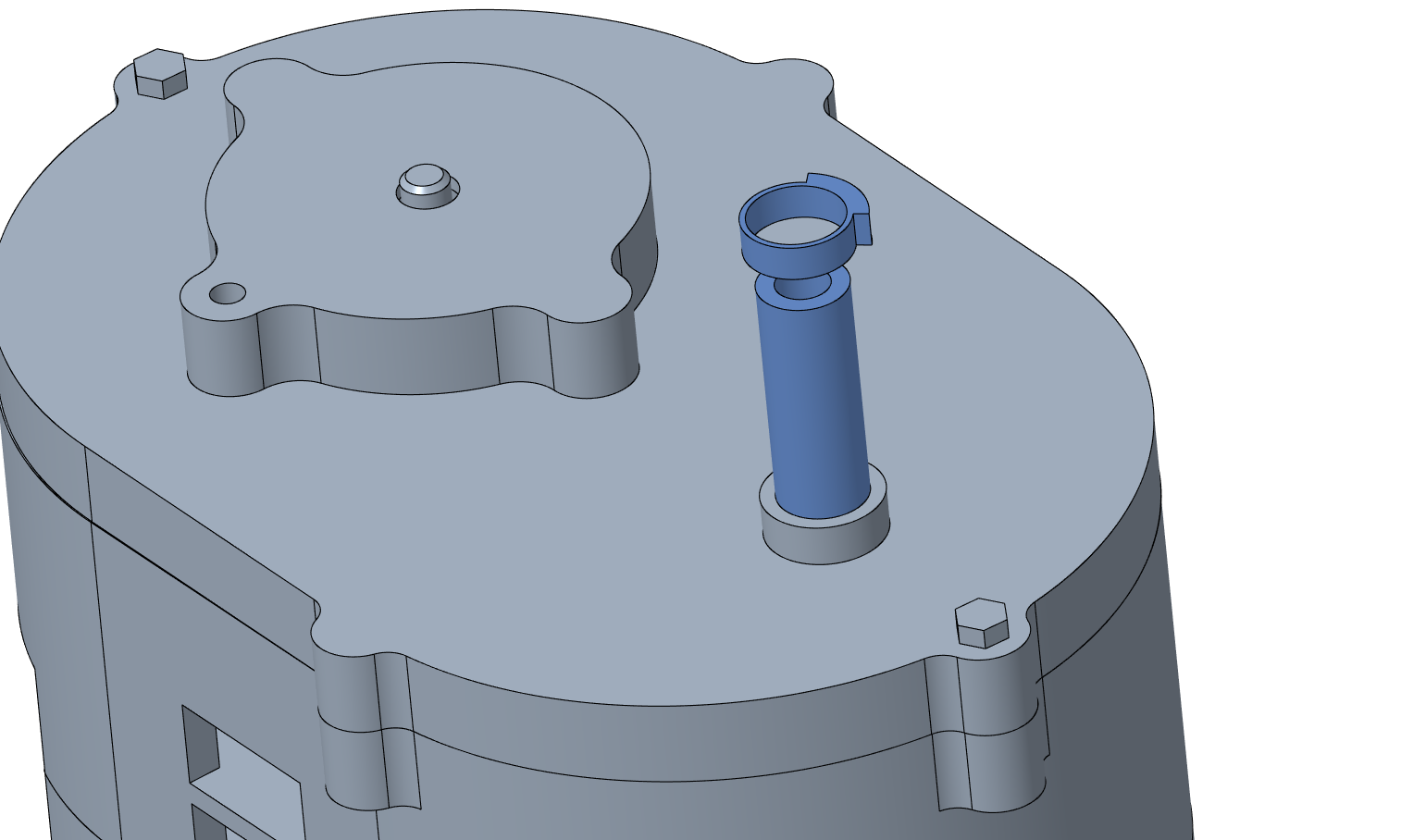}
    \caption{push clamp}
    \label{fig:benchmark_push_clamp}
\end{subfigure}
\begin{subfigure}[t]{0.19\textwidth}
    \captionsetup{justification=centering}
    \includegraphics[width=\textwidth]{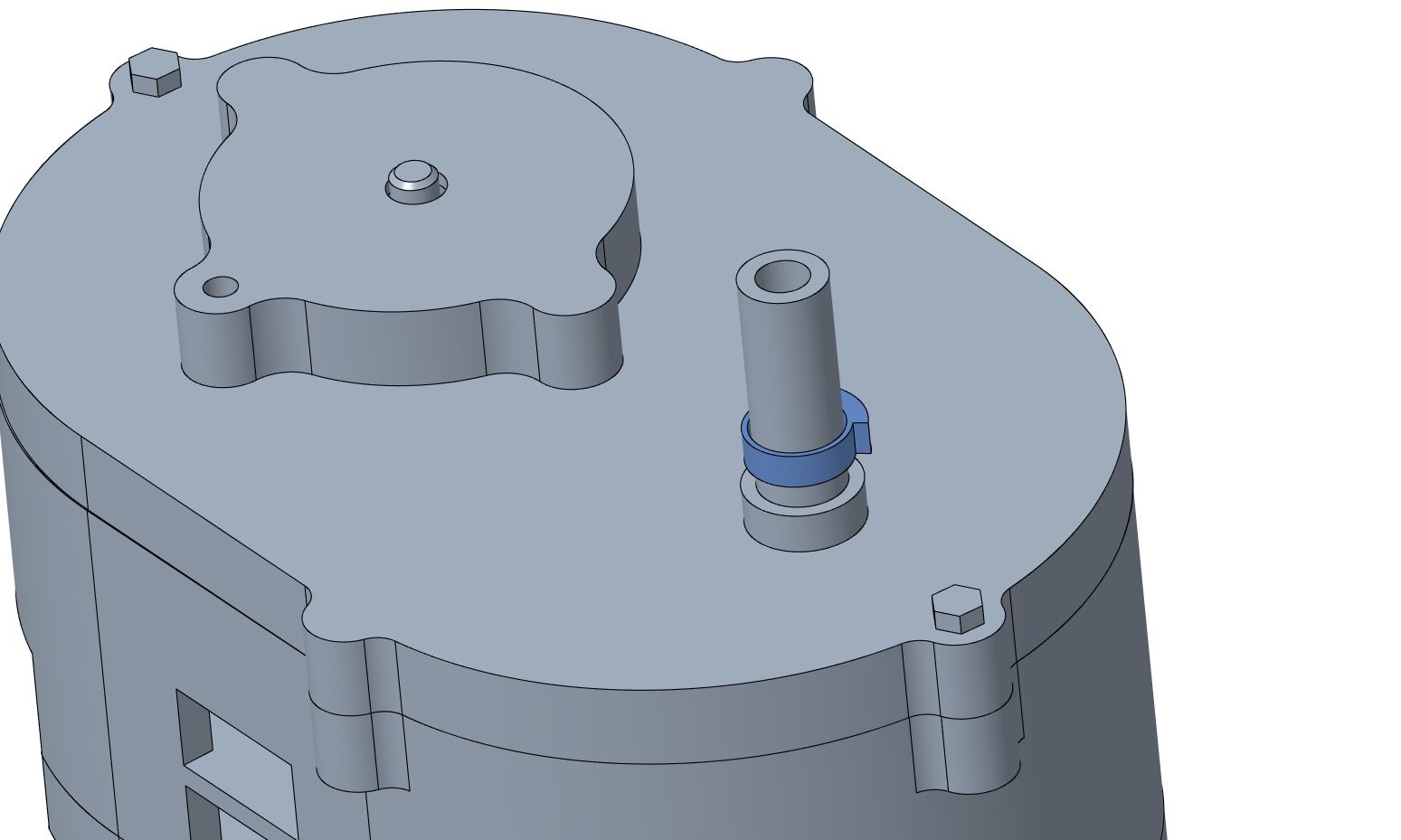}
    \caption{clamp together}
    \label{fig:benchmark_clamp_together}
\end{subfigure}

\caption{Overview of assembly tasks required to successfully assemble the benchmark. Tasks occurring multiple times like peg insertion are only depicted once. Blue colored objects show the interacting components during task execution.}
\label{fig:benchmark_processes_overview}

\end{figure*}
The benchmark is designed to align with the specific demands of industrial assembly and represents industrial processes in a way that can be used for research purposes. A key feature is its stepwise structure, where further assembly is only possible if the preceding step is successfully completed. 

The approach is inspired by the design principles of electric drives, specifically those found in e-bikes.
Building on this inspiration, an assembly is provided that includes 59 modular assembly tasks, integrating mechanical and electronic components.
The 59 primitives span seven operation types: insertion (peg-in-hole, connector, and component insertion), screwing, press fitting, snapping, clamping, reorienting and grasping.
They are distributed across three hierarchically ordered subassemblies: SA1 (bottom housing), SA2 (middle housing module), and SA3 (final integration), each containing a fixed sequence of subtasks that must be completed in order.
For its use in a research environment, the components have a reduced complexity. To preserve the integrity of components and to enable the robot to apply the necessary force during assembly, the process is limited to grasping and joining operations with low-force impacts. 
The key components of the assembly include a rotor, a stator, a simple gearbox, control electronics and associated connectors.

\subsection{Base Parts}
The benchmark incorporates essential base parts and adheres to standardized procedures as defined in \cite{DIN8593-1} and \cite{DIN8593-3}, see Fig. \ref{fig:experimental_setup}.
The processes cover general assembly as well as snapping and placing operations. Further assembly steps and auxiliary processes are integrated into the workflow.

The model is organized into three subassemblies: bottom, middle and top housing. Each subassembly involves steps that adhere to its structural and functional requirements.

The components used include both standard and custom parts. Standard parts such as screws, pins, plugs, o-rings, hoses, and matching hose clamps are employed to ensure compatibility and ease of assembly. In addition, 3D-printed parts are utilized for specific components, including the three housing elements, the rotor, the PCB, two gear wheels and the cover plate. It was decided to 3D print the components, as they possess specific shapes and properties that would otherwise require custom manufacturing. These are exclusively components whose functionality is not compromised by the method of production.
The gears were modeled in accordance with DIN 780 and adapted to the application case. Both gears are of equal size, have 35 teeth, and a module of 2. Instead of the intended inner diameter of 16 mm, the inner diameter was reduced to 8.2 mm to ensure compatibility with the other components. 
For printing with the material PLA, a Bambu Lab X1-Carbon was utilized. The settings are available in the provided documentation.
Due to the utilization of 3D printing technology and standardized components, this represents a cost-effective solution, with expenses amounting to approximately 90€. Furthermore, this approach ensures that the benchmark can be easily replicated.
Several design challenges were encountered. Sourcing internationally available standard components required careful supplier selection; 
the Bill of Materials is restricted to parts accessible through widely available distributors.
Achieving reliable dimensional fits with 3D-printed parts required local CAD tolerancing to compensate for FDM shrinkage;
the final settings and adapted files are provided on the project website. The fixture and provisioning board layout was determined through repeated reachability tests to keep all operations within the robot's workspace.
All relevant documents can be found at {\href{https://irp-hka.github.io/robotic-assembly-benchmark/}{https://irp-hka.github.io/robotic-assembly-benchmark/}}. These include materials such as the evaluation protocol, CAD files of the parts to be printed, the parts list and the assembly instructions.
\definecolor{color_material_soft}{rgb}{0.41708574, 0.68063053, 0.83823145}
\definecolor{color_material_medium}{rgb}{0.12710496, 0.44018454, 0.70749712}
\definecolor{color_material_hard}{rgb}{0.03137255, 0.18823529, 0.41960784}
\newcommand{\sqboxs}{1.2ex}
\newcommand{\sqboxf}{0.6pt}
\newcommand{\sqbox}[1]{\textcolor{#1}{\rule{\sqboxs}{\sqboxs}}}
\newcommand{\sqboxblack}[1]{\setlength{\fboxsep}{0pt}\fbox{\sqbox{#1}}}
\begin{figure}[!t]
    \centering
    \includegraphics[width=\linewidth]{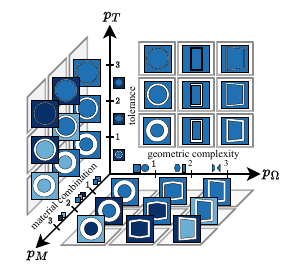}
    \caption{Taxonomy for the categorical assessment of the complexity of robot-guided assembly tasks. The complexity of each assembly task can be evaluated using the three properties tolerance $T$, geometry $\Omega$ and material combination $M$, where $p_i$ encodes the complexity of each group. In the latter, the colors stand for the stiffness of the material, namely \sqboxblack{color_material_soft} for soft, \sqboxblack{color_material_medium} for medium and \sqboxblack{color_material_hard} for hard.}
    \label{fig:taxonomy_3D}
\end{figure}
\subsection{Tasks}
The assembly benchmark begins with the components supplied on a board, each positioned according to the selected planning complexity, as pictured in Fig. \ref{fig:planning_complexity}. 
We provide three levels of planning complexity. Complexity level 1 corresponds to parts being provided by a part feeder.
Complexity level 2 provides the parts unoriented on a plate and complexity level 3 provides the parts cluttered in a box.
These levels govern only the \emph{initial provision} of parts that are individually grasped and inserted, such as screws, pins, plugs, and gear wheels, and do not apply to the housing parts (bottom, middle, and top housing), which are always placed in fixed fixture positions. The downstream assembly sequence and the task-level complexity taxonomy (Section~\ref{ssec:performance_metrics}) remain identical across all three levels.
These three levels of complexity are provided to enable the assessment of a method's ability to handle increasing planning complexity.
For example, certain assemblies can only be created in a designated sequence. Components that are introduced first in the assembly sequence must be found, relocated and grasped first.
This poses a challenge, especially at complexity level 3, which requires more intelligent grasping and planning algorithms.

The bottom housing serves as the foundation, into which all other components are assembled.
Explosion views of exemplary assemblies of the benchmark are illustrated in Fig. \ref{fig:overview_benchmark}.
During assembly, any manual intervention, total time from start to completion, and time for each subassembly group must be documented.

In the following, an overview of the assembly tasks required to assemble the benchmark will be given. 
The tasks are depicted in Fig. \ref{fig:benchmark_processes_overview}, whereby additionally the grasping and aligning of the part in relation to the assembly group is necessary.
However, different requirements for grasping are needed that match the part's structure and its intended space inside the housing.

The printed circuit board needs to be grasped at the socket for the plug. It is then aligned and inserted into the spring-loaded pins (Fig. \ref{fig:benchmark_circuit_board}). Both internal plugs must be grasped from the provisioning board. Proper alignment and insertion need to be ensured. One plug should be connected to the printed circuit board, while the other plug should be connected to the socket in the middle housing (Fig. \ref{fig:benchmark_internal_plug}). The external plug (bayonet connector) is also provided by the provisioning board. When aligning, it is important that the wiring from the plug matches the holes in the socket before the plug is inserted (Fig. \ref{fig:benchmark_external_plug}). The rotor is provided on the board and can be grasped at the tip of its shaft. It must be aligned with the stator in the bottom housing to ensure successful insertion (Fig. \ref{fig:benchmark_rotor}). 

The pegs are each provided on the provisioning board. They must be aligned and inserted into holes that have 0.5 mm clearance (Fig. \ref{fig:benchmark_peg}). 
The center housing section can be grasped in a special grasping notch that is located on the side.
It then has to be rotated by 180° and aligned with the bottom housing.
When aligning, the rotor shaft and both pegs must be aligned with the housing before inserting (Fig. \ref{fig:benchmark_center_housing}). Both gear wheels must be picked from the board.
The gripper can grasp each wheel by its cylindrical shape before aligning and inserting them onto the shafts (Fig. \ref{fig:benchmark_align_gear_wheel} and \ref{fig:benchmark_insert_gear_wheel}). 
The top housing must be aligned with both pegs and placed onto the center housing. 
It can be grasped through the opening from the gear shaft or at the hose fitting (Fig. \ref{fig:benchmark_align_top}). 
The screws are to be picked from the provisioning board before being aligned with the through holes and screwed in (Fig. \ref{fig:benchmark_screwing}). Prior to aligning with the notch, the o-ring needs to be picked from the provisioning board. To ensure easy access with the gripper, half of the ring is freely accessible.
It is then inserted into the notch that surrounds the rotor shaft (Fig. \ref{fig:benchmark_oring}). 

The cover plate is best grasped at the rim before it is aligned with the pegs and placed on the top housing (Fig. \ref{fig:benchmark_align_coverplate}).
When picking the hose, grasping tightly must be avoided to prevent the collapse of the soft material.
It is then placed on the nozzle (Fig. \ref{fig:benchmark_slide_hose}). To stop the hose from sliding off, a clamp is added.
When picking the clamp from the provisioning board, it is grasped softly to avoid closure of the clamp in the wrong position.
Once aligned on the hose, it can be clamped together. The clamp is considered secure when the hose cannot be manually shifted (Fig. \ref{fig:benchmark_push_clamp} and \ref{fig:benchmark_clamp_together}).

The assembled module must be grasped, elevated from its current position, rotated by 180°, and subsequently returned to its original location in the middle housing.
The final two screws must be selected from the provisioning board, ensuring that they are aligned with the through holes. These screws should then be inserted into the bottom housing, thus securing it to the middle housing.
The benchmark has now been fully assembled.

\subsection{Performance Metrics}
\label{ssec:performance_metrics}
The evaluation of individual assembly tasks and subassembly groups is categorized into different complexity levels. 
Each level corresponds to a certain number of points, where more challenging tasks result in more points on successful completion.

\subsubsection{Taxonomy of Assembly Complexity}
We propose a systematic way of addressing assembly complexity by assigning points to the assembly tasks based on the properties of their joining parts.
Depending on the assembly task, there are numerous properties to be considered to describe its complexity. 
To achieve a concise formulation, three principal properties have been identified and applied to all tasks.

\textbf{Tolerance:} The tolerance between the joining parts is a key factor in determining the difficulty of an assembly task.
Depending on the tolerance, more points are assigned if the tolerance is low and fewer if the tolerance is high.
We define the tolerance limits using the repeatability of robots in accordance with ISO 9283 \cite{ISO9283} together with general fit tolerances in accordance with DIN ISO 2768-1 \cite{DINISO2768-1}.
We assume a high repeatability of $\pm 0.03$ mm. For the fit tolerances, we use the medium ($\pm 0.2$ mm), coarse ($\pm 0.5$ mm), and very coarse ($\pm 1$ mm) tolerance classes.
Together, this results in tolerances of $0.46$ mm, $1.06$ mm, and $2.06$ mm, whereby rounding up to one decimal place results in the following point allocation:
\begin{align}
    D - d &\geq  2.1\text{ mm} & p_T &= 1, \\
    1.1 \leq D - d & < 2.1\text{ mm} & p_T &= 2, \\
    D - d &< 1.1\text{ mm} & p_T &= 3,
\end{align}
where $D$ and $d$ are the outer and inner diameter, respectively.

\begin{figure}[t]
    \centering
    \includegraphics{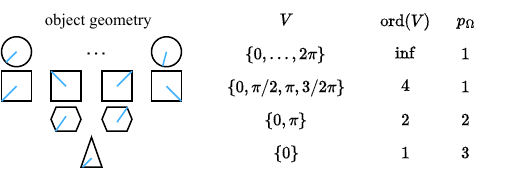}
    \caption{Examples of point allocation for different geometric complexities $\text{ord}(V)$. The blue line indicates the orientation of the object.}
    \label{fig:geometric_complexity}
\end{figure}

\textbf{Geometric complexity:} We define the geometric complexity by the order of symmetry. The number of unique rotations that map the object onto itself is given by the order of the cyclic subgroup $V$.
For example, a circle can be rotated in any way, $V=[0,2\pi )$, around the $z$-axis while still being correctly aligned to the hole.
In contrast, a square yields 4 possible configurations $V= \{ 0, \frac{\pi}{2}, \pi, \frac{3\pi}{2} \}$, therefore being more complex.
See Fig. \ref{fig:geometric_complexity} for examples.
The cyclic subgroup is hereby defined by (\ref{eq:cyclic_subgroup}), where $\Omega \subset \mathbb{R}^2$ is the cross-sectional shape of the object represented as a set of points and $\mathbf{A}$ is the rotation matrix with rotation angle $\theta$.
\begin{align} \label{eq:cyclic_subgroup}
    V = \{\theta \in [0,2\pi) \quad | \quad \mathbf{A}(\theta) \cdot \Omega = \Omega , \quad \mathbf{A} \in SO(2) \}.
\end{align}

Points are assigned based on the complexity of the object, which corresponds to the order of $V$:
\begin{align}
    \text{ord}(V) &\geq 4 & p_{\Omega} &= 1, \\
    4 > \text{ord}(V) &\geq 2 & p_{\Omega} &= 2, \\
    \text{ord}(V) &= 1 & p_{\Omega} &= 3.
\end{align}

\textbf{Material combination:} The difficulty arising from the materials of two mating parts is governed by the  resulting overall stiffness of their contact rather than by the material pairing itself.
Since both parts act as springs in series, the contact is characterized by the effective Young's modulus
\begin{equation} \label{eq:E_eff}
    E_{\mathrm{eff}} = \left( \frac{1}{E_1} + \frac{1}{E_2} \right)^{-1},
\end{equation}
where $E_1$ and $E_2$ are the Young's moduli of the two mating parts.
Because $E_{\mathrm{eff}}$ is dominated by the more compliant part, this single quantity captures two distinct failure mechanisms, resulting in a U-shaped relation between stiffness and difficulty.
For stiff contacts, small pose deviations cause large contact forces, which promotes jamming and wedging~\cite{Zhao_2026}.
For compliant contacts, the parts deform significantly during assembly, so that the rigid-body assumption no longer holds.
Intermediate stiffnesses, where forces remain bounded and parts behave rigidly, are the least demanding.
Accordingly, $E_{\mathrm{eff}}$ is binned on a logarithmic scale and mapped to $p_M$ as listed in Table~\ref{tab:material_complexity}.

\begin{table}[t]
    \centering
    \caption{Point allocation for the material combination based on the effective Young's modulus $E_{\mathrm{eff}}$ according to (\ref{eq:E_eff}).}
    \label{tab:material_complexity}
\resizebox{\linewidth}{!}{%
\begin{tabular}{@{}lcll@{}}
\toprule
$E_{\mathrm{eff}}$ in GPa & $p_M$ & Regime & Typical pairings \\ \midrule
$< 0.1$       & 3 & Large deformation           & Elastomer involved (o-ring, tube) \\
$0.1$ -- $1$  & 2 & Noticeable deformation      & Soft thermoplastics (PE, PTFE) \\
$1$ -- $10$   & 1 & Rigid, force-tolerant       & Engineering polymers, polymer--metal \\
$10$ -- $100$ & 2 & Increased force sensitivity & Light metals (Al--Al, Al--steel) \\
$\geq 100$    & 3 & High force sensitivity      & Steel--steel, ceramics \\ \bottomrule
\end{tabular}}
\end{table}

\begin{figure}[t]
    \centering
    \includegraphics[width=\linewidth]{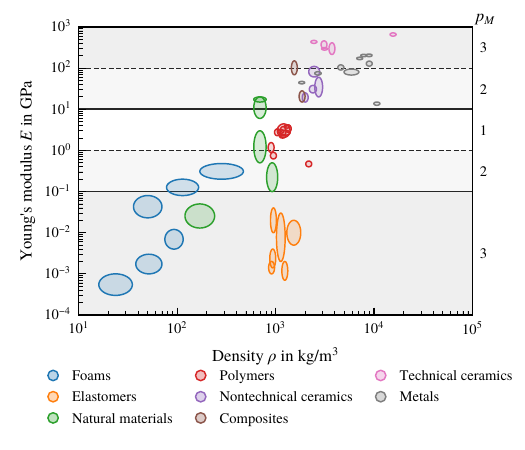}
    \caption{Young's modulus over density for families of engineering materials (after~\cite{ashby2011}) with the bin edges of the material combination. Solid lines mark the primary edges placed in the gaps between the elastomer, polymer and metal families; dashed lines mark the secondary edges obtained by decade subdivision. The resulting points $p_M$ are given on the right axis.}
    \label{fig:ashby}
\end{figure}

The bin edges are not fitted to the benchmark.
The two primary edges at 0.1 GPa and 10 GPa are placed in the two principal gaps of the modulus spectrum of engineering materials, between the elastomer and polymer families and between the polymer and metal families, respectively, as shown in Fig. \ref{fig:ashby} \cite{ashby2011}.
Each outer region is further subdivided by one decade, at 1 GPa and 100 GPa, to distinguish moderate from extreme behavior.
Foams, natural materials and fiber composites, for which these family gaps do not exist, are scored by their $E_{\mathrm{eff}}$ in the same manner.

The 1--3 point ranges for each property were chosen as the minimal integer scale that yields meaningful differentiation while remaining practically applicable across diverse assemblies. These values represent initial reference values intended to anchor community-wide comparisons; researchers applying the taxonomy to other benchmarks are encouraged to refine the thresholds based on accumulated data and to report any adapted values explicitly so that scores remain comparable.
Fig. \ref{fig:taxonomy_3D} visualizes the taxonomy.
The total points of each primitive $i$ are then calculated as the sum of points for each property, where $K$ is the total number of properties and $k \in \{T,\Omega,M \}$. 
\begin{equation}
    P_i = \sum_{k}^K p_{ik} = p_{iT} + p_{i\Omega} + p_{iM},
\end{equation}

\begin{table}[t]
\centering
\caption{Overview of the metrics used in this work. The table defines symbols for complexity points ($p_{ik}$), reliability measures ($R_i$, $R_{sa}$, $R_c$, $R$), scores ($S_i$, $S_{sa}$, $S_c$, $S$), timing metrics ($T_{sa}$, $T$), and force interaction metrics ($E_{xy}$, $E_z$, $\sigma_{xy}$, $\sigma_z$).}
\label{tab:nomenclature}
\begin{tabular}{@{}lll@{}}
\toprule
Symbol                  & Meaning                               &  \\ \midrule
$p_{ik}$                & complexity points of property $k$ of primitive $i$ &  \\
$P_i$                   & total points of primitive $i$             &  \\
$R_i$                   & reliability of primitive $i$             &  \\
$R_{task}$      & reliability of subtask ($= \prod_{i \in task} R_i$) &  \\
$R_{sa}$                & reliability of subassembly            &  \\
$R_c$                   & reliability of primitive category     &  \\
$R$                     & reliability of the overall assembly   &  \\
$S_i$                   & score of primitive $i$                    &  \\
$S_{task}$      & score of subtask ($= \sum_{i \in task} S_i$) &  \\
$S_{sa}$                & score of subassembly                  &  \\
$S_c$                   & score of primitive category           &  \\
$S$                     & overall benchmark score               &  \\
$T_{sa}$                & subassembly time                      &  \\
$T$                     & overall assembly time                 &  \\
$E_{xy}$, $E_z$         & force energy                          &  \\
$\sigma_{xy}$, $\sigma_z$ & force smoothness                    &  \\ \bottomrule
\end{tabular}
\end{table}

\subsubsection{Evaluation Protocol} \label{sssec:evaluation_protocol}
Points are assigned based on the taxonomy proposed in this work, where the tuple \{$p_T$, $p_{\Omega}$, $p_M$\} characterizes the difficulty of the respective primitive.
For this benchmark, the difficulty \{$p_{iT}$, $p_{i\Omega}$, $p_{iM}$\} of each primitive $i$ is predefined and given in the supplementary material.
When two primitives share the same total score $P_i$ but differ in their sub-term profiles, the tuple $\{p_T, p_{\Omega}, p_M\}$ and the corresponding per-category reliabilities $R_c$ should be used to identify which axis drives difficulty.
However, we encourage other researchers to use this taxonomy for their specific robot assembly setup in order to classify assembly tasks uniformly in terms of their complexity and difficulty. This also helps to classify results more accurately across different benchmarks. Furthermore, the overall benchmark score, obtained by summing the taxonomy points across all primitives of a benchmark, serves as a complexity indicator that allows entire benchmarks to be compared with one another in terms of their overall task difficulty.
To evaluate the reliability of the approach, benchmark tests are conducted $N$ times, assessing both the overall assembly performance and the reliability of individual assembly tasks. 

This analysis provides insights into the strengths and weaknesses of a given approach, for instance, excelling in insertion tasks while exhibiting lower performance in screw fastening. 
The following paragraph explains the calculation of the metrics and overall point allocation. An overview of the used metrics is given in Table~\ref{tab:nomenclature}.

For each primitive $i$, the reliability $R_i$ is calculated by (\ref{eq:reliability}), where $b_{ij}$ is a binary value determining if repetition $j$ of primitive $i$ was successful.
\begin{equation} \label{eq:reliability}
R_i = \frac{1}{N} \sum_{j=1}^{N} b_{ij} \in [0, 1].
\end{equation}
The reliability $R_{sa}$ of successfully assembling a specific subassembly is then given by the product law of reliability, where $m$ is the total number of primitives required to complete the subassembly:
\begin{equation}
    R_{sa} = \prod_{i}^m R_i.
\end{equation}

To enable a more detailed evaluation, we also calculate the overall reliability for each primitive category $R_c$.
In this benchmark, the primitive category corresponds directly to the primitive type: \texttt{grasp}, \texttt{reorient}, \texttt{insert}, \texttt{screw}, \texttt{snap}, \texttt{press\_fitting}, and \texttt{clamp}.
The reliability of the overall assembly $R$ is given by the product of the reliabilities of all subassemblies.
The score of each primitive is calculated by:
\begin{equation}
S_i = P_i \cdot R_i.
\end{equation}
A subtask groups one or more primitives (e.g., \texttt{grasp}, \texttt{move}, \texttt{insert}); its score $S_{task}$ is the sum of the scores of its constituent primitives:
\begin{equation}
    S_{task} = \sum_{i \in task} S_i.
\end{equation}
The score for each subassembly $S_{sa}$ is therefore given by the sum of all its subtask scores.
As before, we also calculate the score for each primitive category $S_c$.

The overall benchmark score is then calculated by:
\begin{equation}
   S = \sum_{i=1}^I S_{task,i},
\end{equation}
where $I$ is the total number of assembly subtasks executed to complete the benchmark.
\begin{figure}
    \centering
    \includegraphics[width=.98\linewidth]{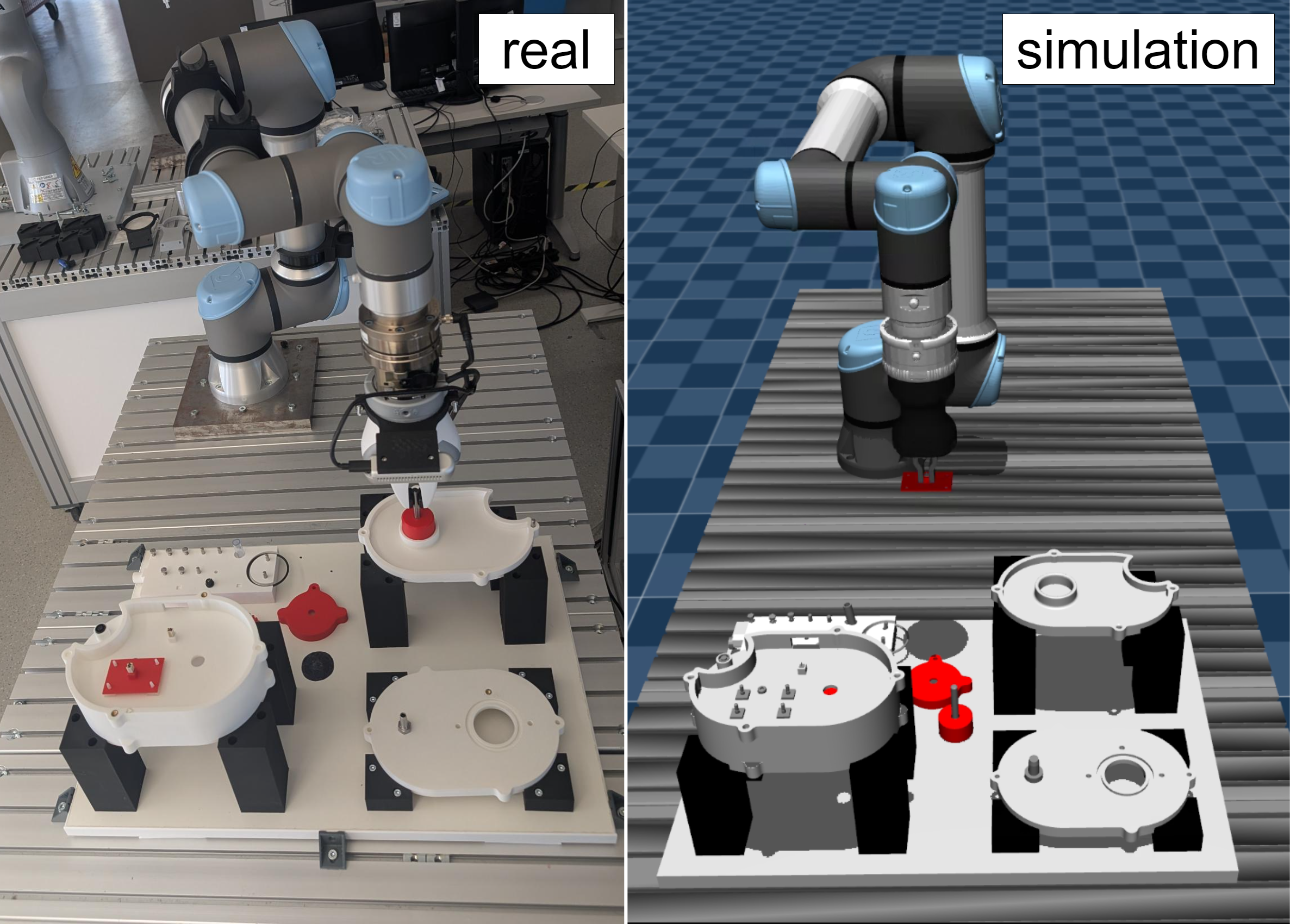}
    \caption{Benchmark setup in real environment (left) and simulation (right), showing the robot, gripper, wrist camera and force-torque sensor. The parts are provided by the part feeder, corresponding to planning complexity level 1.}
    \label{fig:real_sim_setup}
\end{figure}
Additionally, the time taken for each subassembly $T_{sa}$ and the final assembly $T$ are monitored and analyzed. The average completion time is hereby defined by (\ref{eq:average_completion_time}), where $t_{ij}$ is the time required to complete repetition $j$ of subassembly $i$.
\begin{equation} \label{eq:average_completion_time}
T_{sa} = \frac{1}{N} \sum_{j=1}^{N} t_{ij},
\end{equation}
The total completion time is then calculated as the sum of all subassembly completion times.

In addition to task-level metrics, we also use metrics to evaluate the interaction forces.
We calculate the force energy and force smoothness as proposed by \cite{SchemppZhangFriedrichHein2025} for each primitive repetition,
where the force energy $E$ is defined as the cumulative force applied over the course of the insertion process:
\begin{equation}
    E = \frac{1}{N} \sum_{n=0}^{N} |\mathbf{F}(n)|^2,
\end{equation}
where $\mathbf{F} = \mathbf{F}_z$ to calculate the energy $E_z$ in the direction of insertion and $\mathbf{F} = \|\mathbf{F}_x, \mathbf{F}_y\|$ to calculate the energy $E_{xy}$ of the incurred forces on the plane orthogonal to the insertion direction.
Force smoothness $\sigma_{xy}$ or $\sigma_z$, depending on the force component, is defined as the standard deviation of the force signal’s rate of change:
\begin{equation}
    \sigma = \sqrt{ \frac{\sum(\mathbf{\dot{F}} - \mu)^2}{N}},
\end{equation}
where $\mu$ denotes the mean of $\mathbf{\dot{F}}$ over the measurement window.
As before, we distinguish between the direction of insertion $\mathbf{\dot{F}} = \mathbf{\dot{F}_z}$ and the plane orthogonal to it
$\mathbf{\dot{F}} = \mathbf{||\dot{F}_x,\dot{F}_y||}$.
The mean and standard deviation are shown by boxplots.
This analysis enables the identification of subtasks that generate high interaction forces and those that exhibit significant fluctuations.
This observation may suggest potential instability in the approach under investigation.

\begin{figure}
    \centering
    \includegraphics[width=\linewidth]{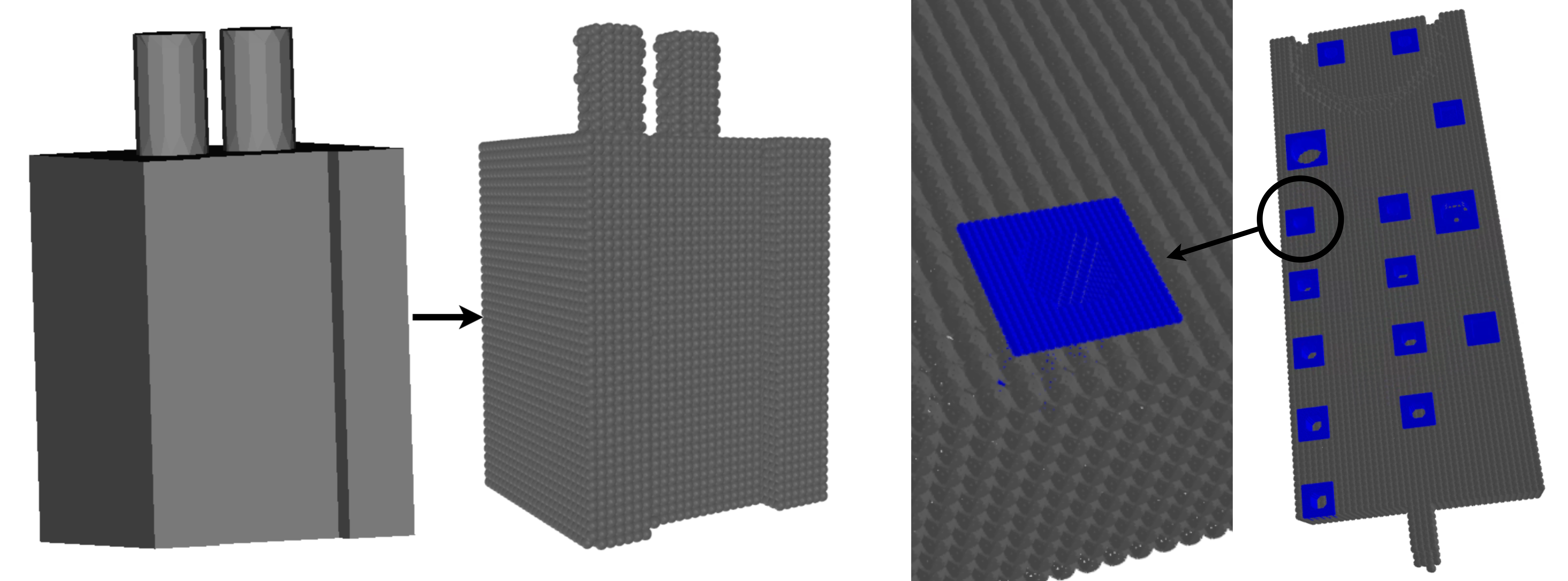}
    \caption{Convex decomposition of assembly parts for accurate contact calculation in simulation.
    Areas where contact is expected during assembly are modeled with higher detail by using smaller sphere radii (right).}
    \label{fig:convex_decomposition}
\end{figure}

\begin{figure*}[!htbp]
    \centering
    \includegraphics[width=\linewidth]{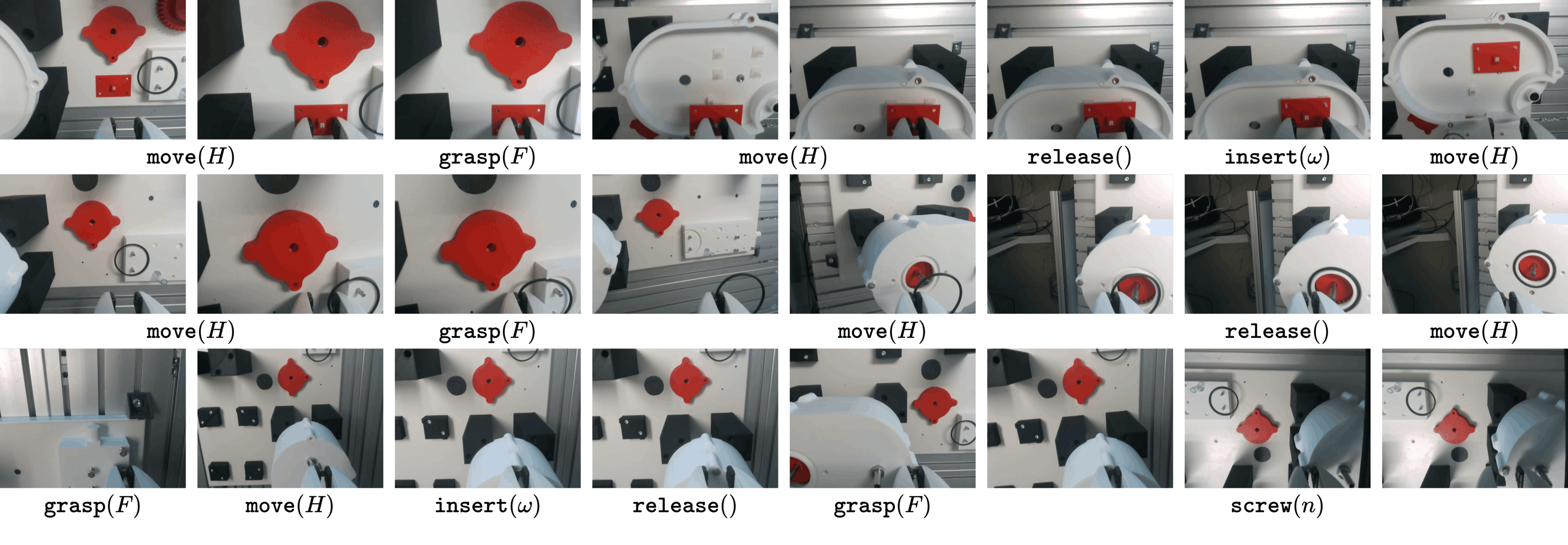}
    \caption{Visualization of executed primitive sequences as seen by the wrist-mounted camera during baseline execution. Each row shows consecutive frames spanning the approach, contact, and completion phase of a single primitive. Top row: PCB insertion, where the board is grasped at its connector socket, aligned above the spring-loaded pins, and pushed into its seat. Center row: o-ring placement, where the ring is picked from the provisioning board, transported to the housing notch, and seated around the rotor shaft. Bottom row: screwing, where the screw is grasped, aligned with the through hole, and driven in by rotating the end effector about the assembly axis.}
    \label{fig:baseline}
\end{figure*}
The overall score $S$ serves as the \emph{primary} headline metric for ranking approaches on LABIT, as it jointly captures both task difficulty (through $P_i$) and execution reliability (through $R_i$). The timing metric $T$ and force-interaction metrics $E_{xy}$, $E_z$, $\sigma_{xy}$, $\sigma_z$ are \emph{secondary, diagnostic} measures that characterize efficiency and contact quality but are not used for ranking.
The allocation of points on a per-subtask basis permits a comprehensive assessment of the respective approach. Consequently, this enables the method's strengths and weaknesses to be identified in a targeted manner.
Building on the point-based performance evaluation and time monitoring, further research can be undertaken, enabling comparative analysis. 
\begin{table}[t]
\centering
\caption{MuJoCo simulation parameters used in LABIT.}
\label{tab:sim_params}
\resizebox{\linewidth}{!}{%
\begin{tabular}{@{}ll@{}}
\toprule
\textbf{Parameter} & \textbf{Value} \\ \midrule
Timestep            & 0.5\,ms (2\,kHz) \\
Integrator          & \texttt{implicitfast} \\
Solver              & Newton \\
Solver tolerance    & $10^{-10}$ \\
Gravity             & $[0,\,0,\,-9.81]$\,m/s$^2$ \\
Friction cone       & elliptic \\
Impedance ratio (\texttt{impratio}) & 10 \\
material plastic & solref = [0.016, 1.0], density = 1240 \\& sliding friction = 0.31 \\
material steel & solref = [0.001, 1.0], density = 7850 \\& sliding friction = 0.42 \\
FTS sample rate     & 2\,kHz (every sim step) \\
IK solver           & \texttt{trac-ik} \\ \bottomrule
\end{tabular}}
\end{table}
\begin{figure*}[!t]
    \centering
    \begin{subfigure}{\textwidth}
        \centering
        \includegraphics[width=\textwidth]{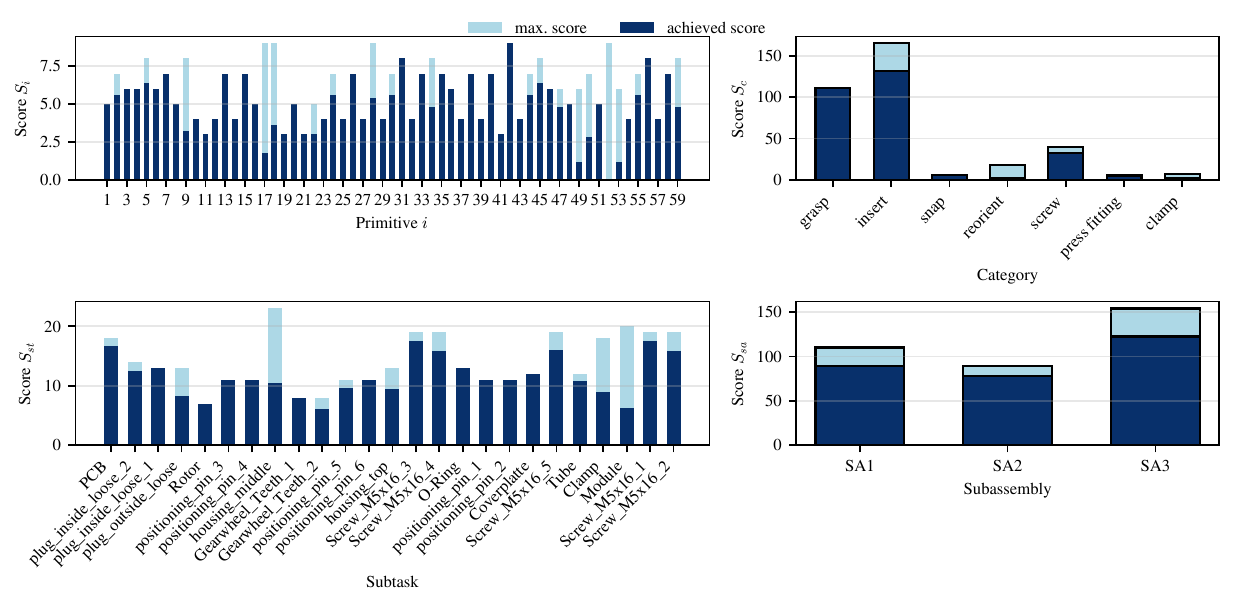}
    \end{subfigure}
    \begin{subfigure}{\textwidth}
        \centering
        \includegraphics[width=\textwidth]{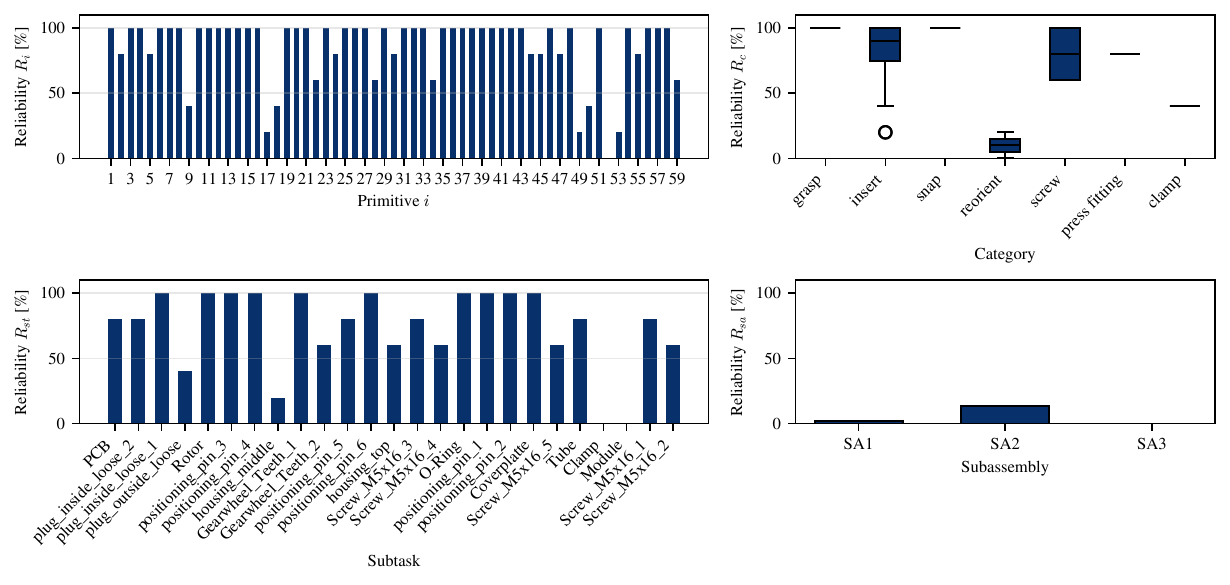}
    \end{subfigure}
    \caption{Results of the baseline implementation using the task-level metrics from Section~\ref{sssec:evaluation_protocol}.
    Top: Scores achieved for each primitive, assembly category and subassembly. Bottom: Reliability for each primitive, assembly category and subassembly.}
    \label{fig:baseline_results}
\end{figure*}

\subsection{Simulation}
LABIT provides a complete simulation environment for physically realistic assembly policy testing, see Fig. \ref{fig:real_sim_setup}.
The environment is made using MuJoCo \cite{todorov2012mujoco}, which provides efficient rigid-body dynamics simulation. The simulation includes accurate collision detection and force computation.
To handle complex geometries in the simulation environment, we employ convex decomposition techniques \cite{SchemppZhangFriedrichHein2025}.
All task objects use sphere-based decomposition (see Fig.~\ref{fig:convex_decomposition}). Sphere radii are chosen per object: smaller objects and areas of expected contact are decomposed in greater detail to ensure accurate force computation.
The simulation environment includes comprehensive logging capabilities for force-torque data, end-effector poses, and joint states during assembly execution.
Additionally, the evaluation protocol is automatically generated from simulation runs, streamlining the assessment process and ensuring consistency across evaluations.
This integration enables researchers to directly compare simulation results with real-world baseline performance.
Table~\ref{tab:sim_params} lists the key simulation parameters used.

\section{Baseline Implementation}
The baseline method employs a classical approach, utilizing manually engineered force-position primitives to execute the assembly plan.
The initial positions of the parts in CAD were registered with the positions on the real benchmark.

\subsection{Primitives}
The primitives employed in the baseline implementation can be categorized into two groups: motion primitives and assembly primitives.
\begin{itemize}
    \item motion primitives:
    \begin{itemize}
        \item \texttt{move($H$)}: move the end effector to pose $H = \begin{bmatrix}
            \mathbf{A} & t \\
            0 & 1
        \end{bmatrix}$, with $\mathbf{A}\in SO(3)$ and $t\in\mathbb{R}^3$.
        \item \texttt{grasp($F$)}: grasp a part with the gripper using grip force $F$.
        \item \texttt{reorient($H$)}: reorient the grasped part to a target pose $H$ in preparation for assembly.
        \item \texttt{release()}: release a part from the gripper.
    \end{itemize}
    \item assembly primitives:
    \begin{itemize}
        \item \texttt{insert($\omega$)}: insert current object in hand using target wrench $\omega \in \mathbb{R}^6$.
        \item \texttt{screw($n$)}: screw current object in hand in target through hole with $n$ rotations.
        \item \texttt{snap()}: press current object in hand into a snap-fit connection.
        \item \texttt{press\_fitting($\omega$)}: press-fit current object in hand into target using wrench $\omega \in \mathbb{R}^6$.
        \item \texttt{clamp($d$)}: clamp the object in hand by setting the distance between the gripper jaws to $d$.
    \end{itemize}
\end{itemize}

\texttt{move} and \texttt{release} are transitional operations and are not individually evaluated. The remaining primitives are invoked across the full benchmark execution as follows: \texttt{grasp} (25), \texttt{reorient} (2), \texttt{insert} (24), \texttt{screw} (5), \texttt{snap} (1), \texttt{press\_fitting} (1), \texttt{clamp} (1).

\begin{figure*}[!t]
    \centering
    \includegraphics[width=\textwidth]{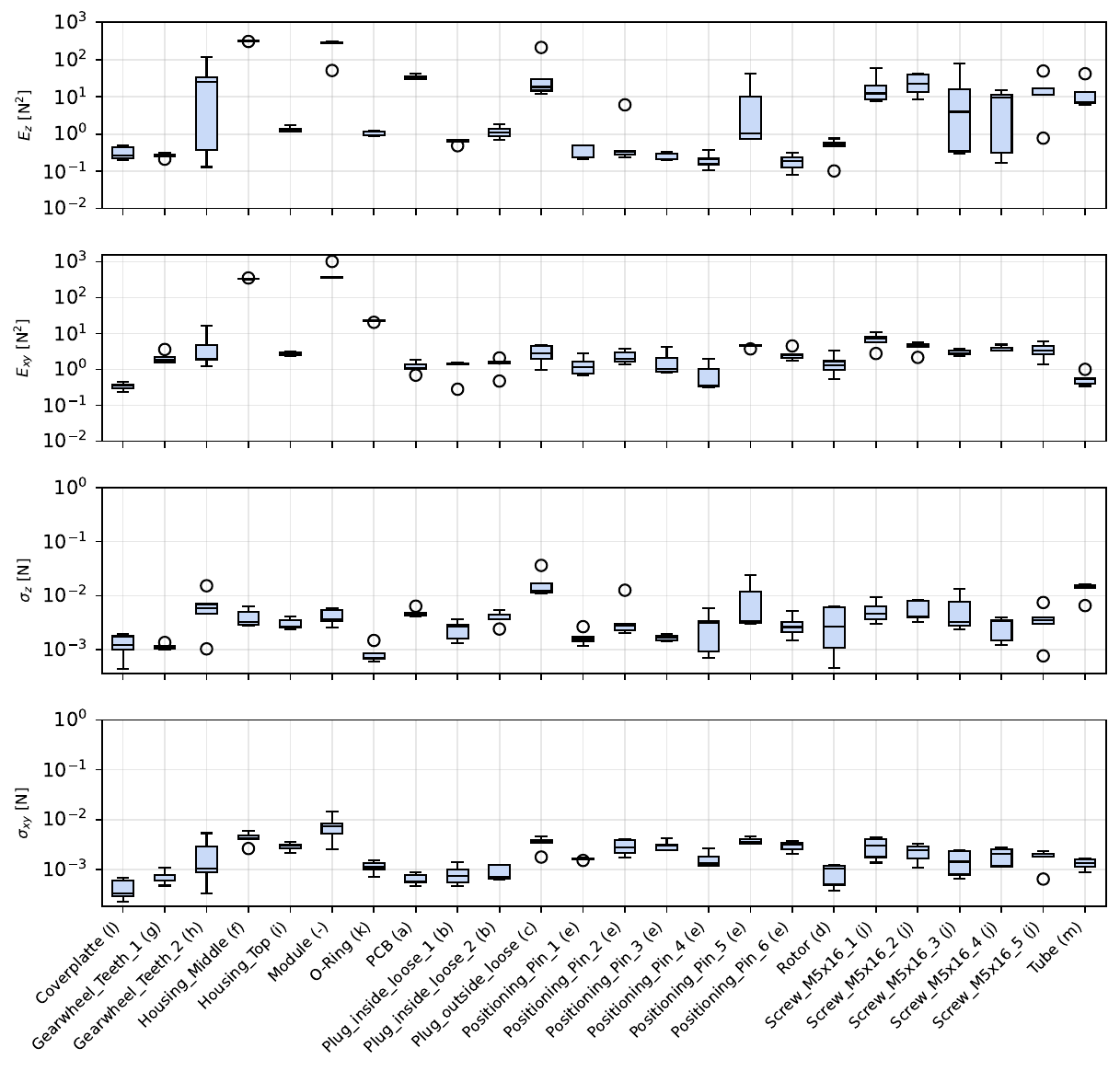}
    \caption{Force interaction metrics for each subtask executed by the baseline method in the real environment.
    The first and second rows show the force energy in $z$-direction and $xy$-direction and the third and fourth rows show the force smoothness in $z$-direction and $xy$-direction respectively.
    The letter in parentheses after the component name refers to the visual representation of the subtask in Fig. \ref{fig:benchmark_processes_overview}.}
    \label{fig:baseline_force_metrics}
\end{figure*}
Based on Finkemeyer \cite{Finkemeyer_2004}, we combine motion and assembly primitives, where each motion primitive may be executed multiple times:
\texttt{move()} $\xrightarrow{}$ \texttt{grasp()} $\xrightarrow{}$ \texttt{move()} $\xrightarrow{}$ \{\texttt{assembly primitive}\} $\xrightarrow{}$ \texttt{release()} $\xrightarrow{}$ \texttt{move()}.
The baseline employs the ROS 2 (Robot Operating System) framework \cite{ros2}, utilizing the ros2\_control library to regulate the robot's motion.
It encompasses fundamental primitives such as movement, grasping and releasing an object, Cartesian space admittance control, and force feedback, which collectively facilitate the execution of contact-rich assembly tasks. For example, the \texttt{insert()} action guides the insertion by a desired target force in the assembly direction. \texttt{screw()} rotates the end effector around the assembly axis while maintaining contact.
The \texttt{clamp()} action brings the gripper jaws closer together while having the object in hand, pressing it together and completing the clamping.
Fig. \ref{fig:baseline} shows assembly sequences as seen by the mounted wrist camera, illustrating the approach, contact, and completion phases 
for three representative primitives: PCB insertion, o-ring placement, and screwing. The close-up perspective reveals the fine alignment required at the
moment of contact, and highlights how even small offsets at the approach phase can propagate into jamming or misalignment during insertion or screwing.

\begin{figure*}[!t]
    \centering
    \begin{subfigure}{0.32\textwidth}
        \centering
        \captionsetup{justification=centering}
        \includegraphics[width=\textwidth]{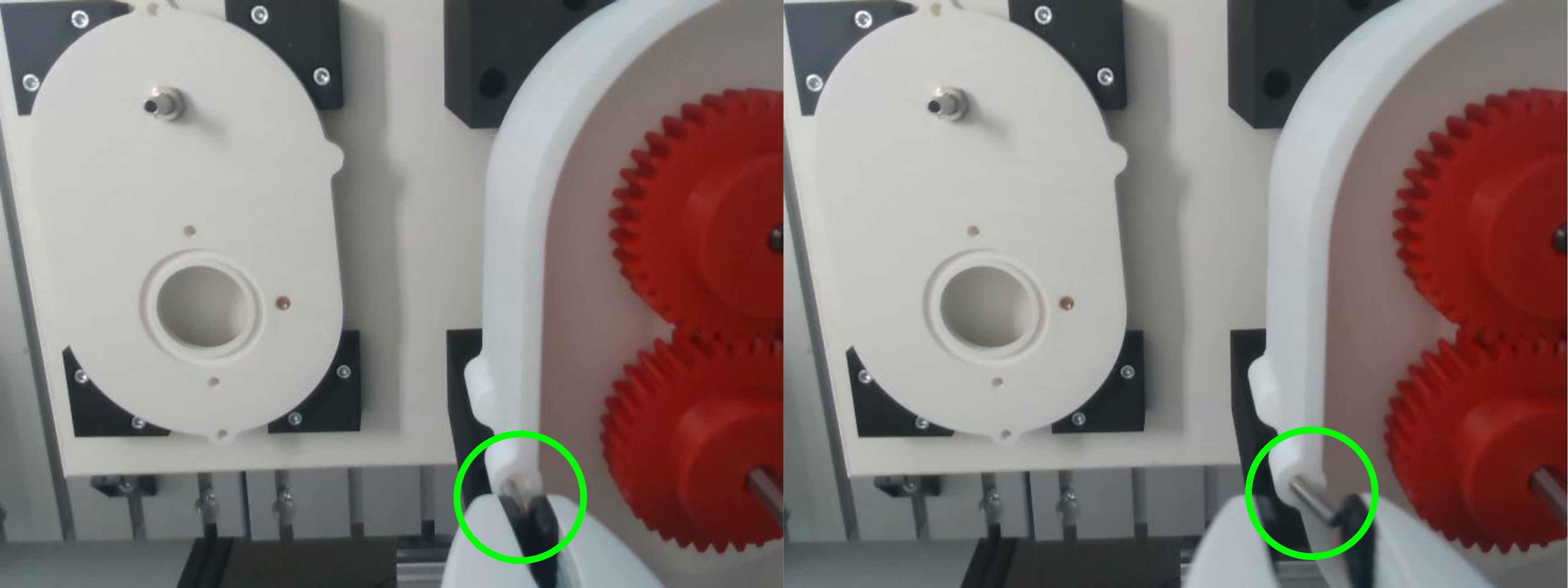}
        \caption{pin insertion failure}
    \end{subfigure}
    \begin{subfigure}{0.32\textwidth}
        \centering
        \captionsetup{justification=centering}
        \includegraphics[width=\textwidth]{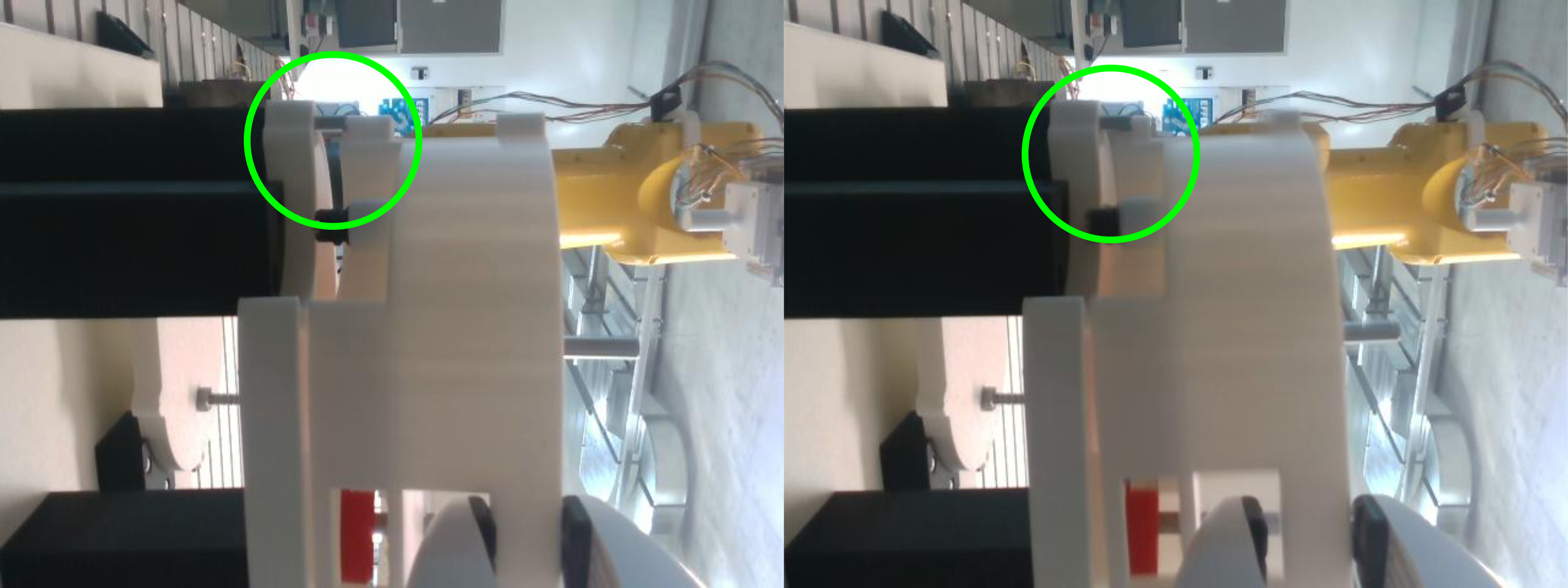}
        \caption{housing middle alignment failure}
    \end{subfigure}
    \begin{subfigure}{0.32\textwidth}
        \centering
        \captionsetup{justification=centering}
        \includegraphics[width=\textwidth]{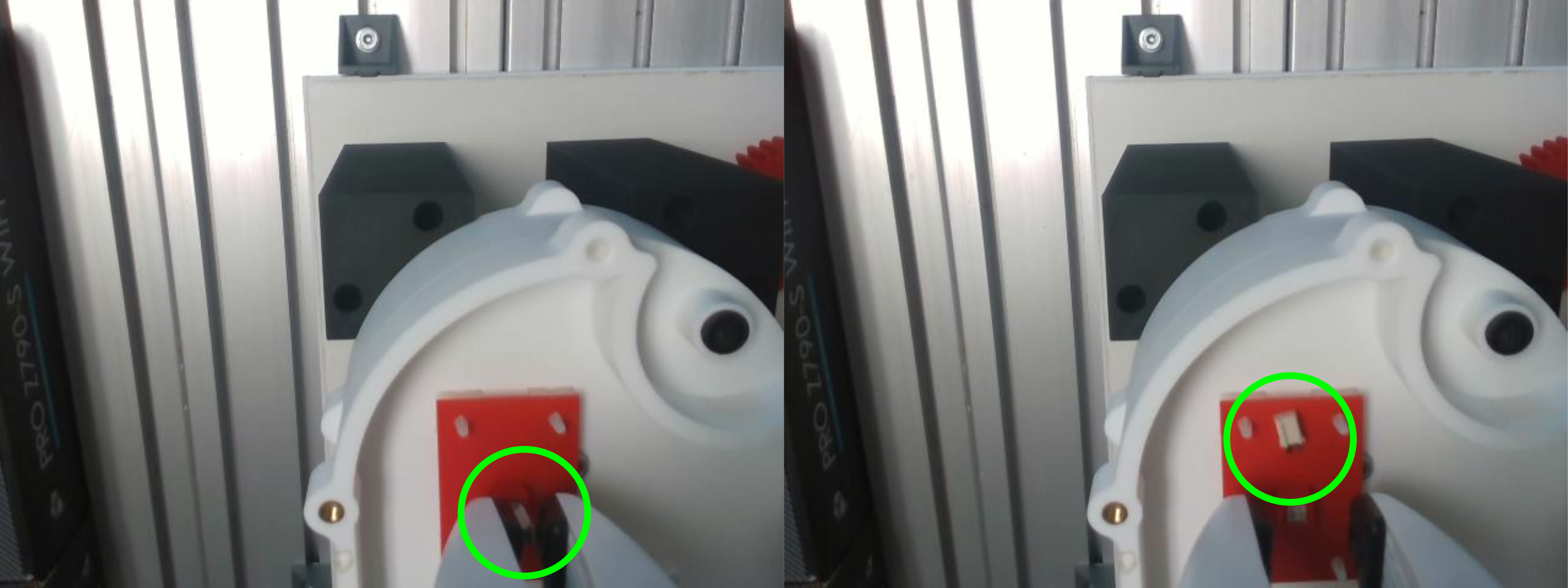}
        \caption{plug insertion failure}
    \end{subfigure}

    \begin{subfigure}{0.32\textwidth}
        \centering
        \captionsetup{justification=centering}
        \includegraphics[width=\textwidth]{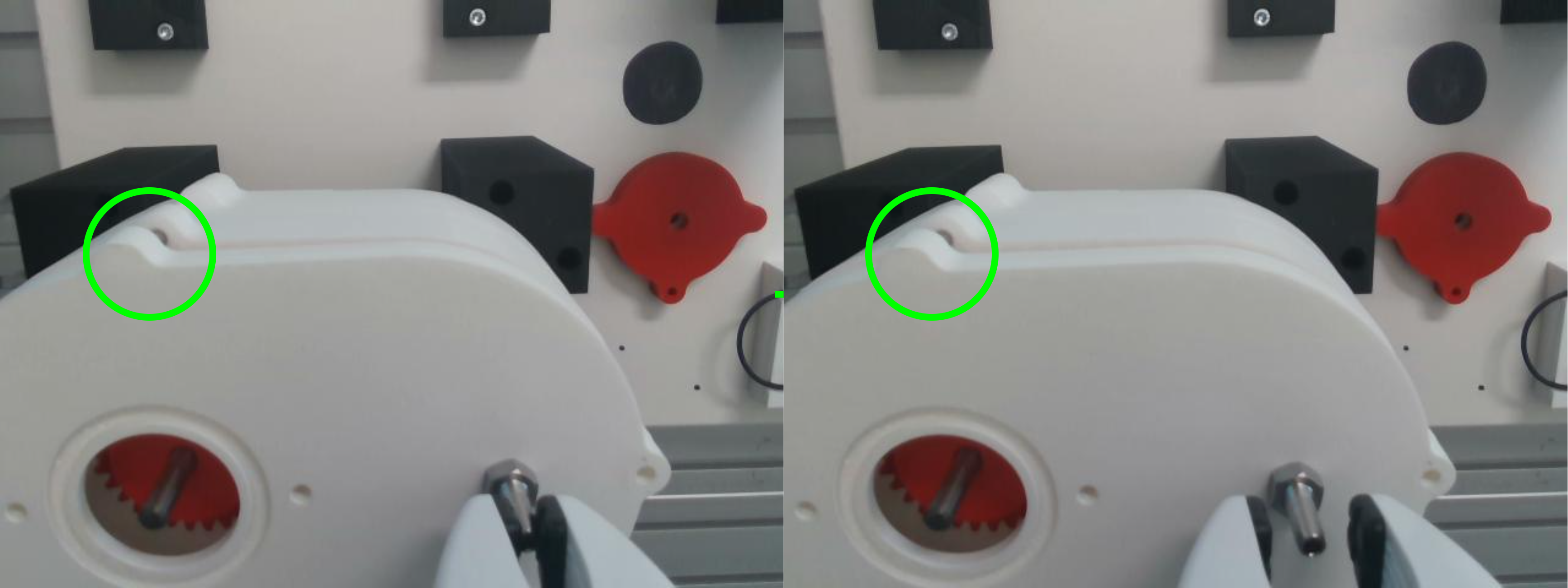}
        \caption{housing top placement failure}
    \end{subfigure}
    \begin{subfigure}{0.32\textwidth}
        \centering
        \captionsetup{justification=centering}
        \includegraphics[width=\textwidth]{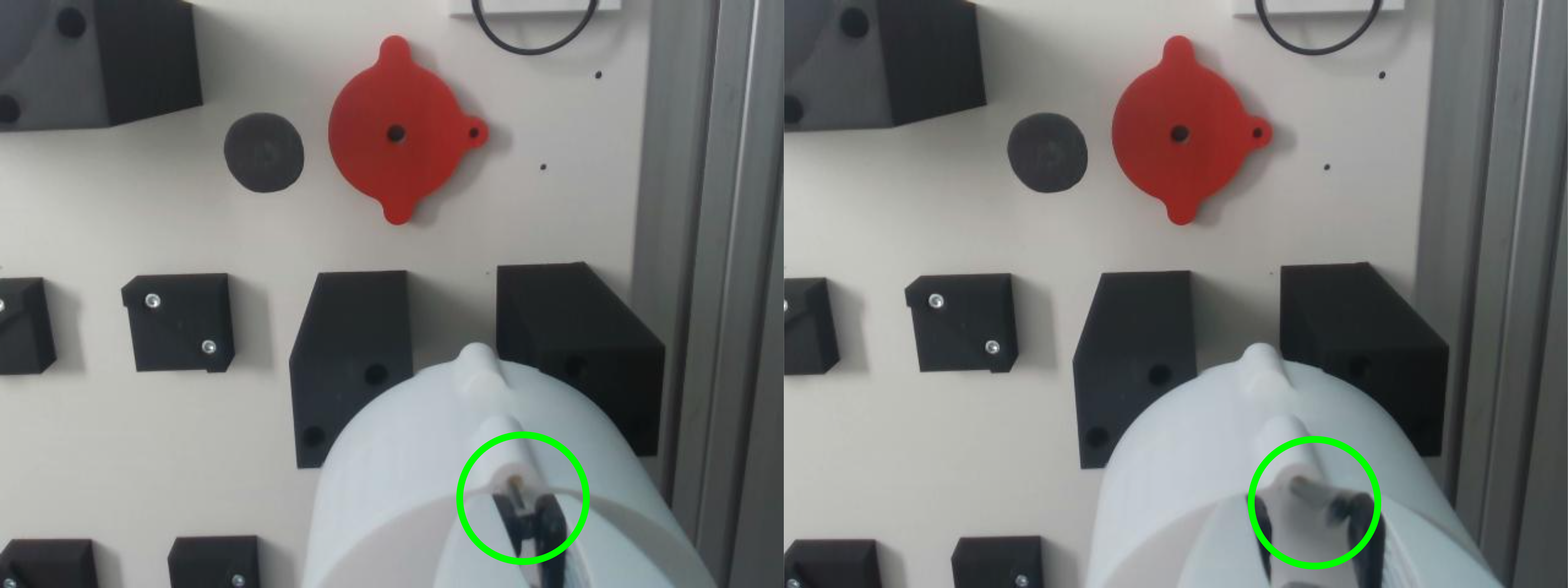}
        \caption{screw insertion failure}
    \end{subfigure}
    \begin{subfigure}{0.32\textwidth}
        \centering
        \captionsetup{justification=centering}
        \includegraphics[width=\textwidth]{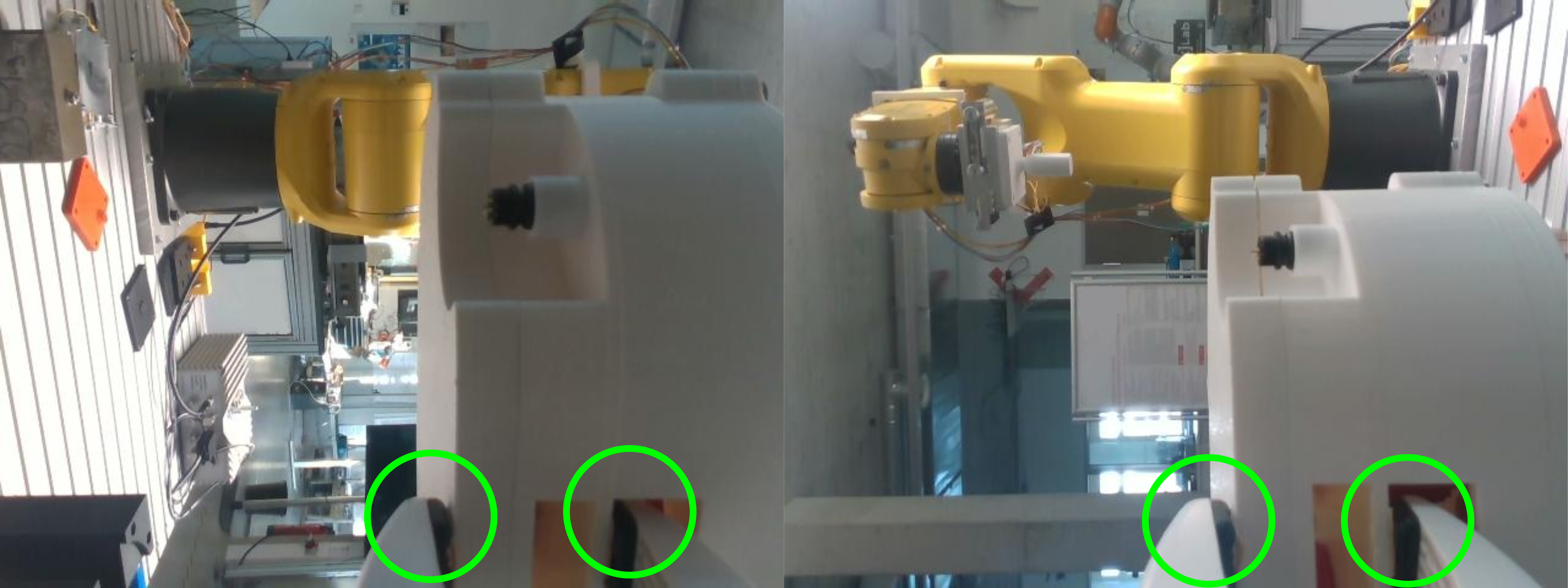}
        \caption{module rotation failure}
    \end{subfigure}
    \caption{Example failure cases during execution of the baseline. (a) inserting a positioning pin fails due to jamming, (b) misalignment of the middle housing during placement, (c) insertion of the internal plug fails due to misalignment, (d) placement of the top housing fails due to slipping of the housing in the gripper, (e) screwing fails due to misalignment of the screw with the through hole and (f) reorienting and placing the module fails due to slipping of the housing in the gripper.}
    \label{fig:failure_modes}
\end{figure*}

\subsection{Results}
We evaluated the baseline method using the taxonomy and evaluation protocol from Section~\ref{ssec:performance_metrics}.
The baseline was executed a total of five times and the results were recorded in the evaluation protocol.
Videos and corresponding position and force-torque data of the baseline execution are available on the website.
The time taken to complete the entire assembly was on average 32 min, with subassembly 1 (SA1) taking 9.8 min, subassembly 2 (SA2) taking 9.7 min, and subassembly 3 (SA3) taking 12.5 min.
The long execution time is mainly due to the low robot velocity used for safety reasons in the real environment.

\textbf{Reliability evaluation:}
Fig. \ref{fig:baseline_results} shows the achieved scores and reliabilities of the baseline method for each primitive, assembly category and subassembly.
The \texttt{grasp} category achieved a reliability of 100\%, as all parts were successfully grasped in each repetition.
The high reliability can be explained by the use of a part feeder, providing consistent part positions. Using planning complexity level 2 or 3 (unoriented plate or cluttered box) would likely lead to lower reliabilities in this category.
The \texttt{insert} category achieved a mean reliability of 82.3\%. Outliers include the reorientation and alignment of the housing module with a reliability of 0\%, reorienting and attaching the middle housing to the bottom housing with a reliability of 20\%,
and inserting the outside plug with a reliability of 40\%.
The \texttt{screw} category achieved a mean reliability of 80\%, with the lowest success rate being 60\%.
The most challenging tasks were reorienting and aligning the housing parts. Small deviations in the positions of the housing and assembly parts result in assembly failure due to small tolerances.
Additionally, since the baseline does not use vision as part of its framework, reorientation of the housing part is difficult and often leads to failure, as the housing parts slip inside the gripper.
This results in outliers for the \texttt{insert()} primitive, as can be seen in Fig. \ref{fig:baseline_results}.
A note on the sequential-dependency: individual primitive reliabilities $R_i$ are measured independently across all $N$ runs.
When a primitive fails, the assembly state is manually corrected before the next primitive is attempted, so that downstream steps can still be evaluated independently.
A 0\% individual reliability therefore does not prevent subsequent primitives from being executed and measured.
The cascading failure effect of the strict sequential dependency is captured at the subassembly level through $R_{sa} = \prod_i R_i$, which will equal zero whenever any constituent $R_i = 0$.
The housing-module reorientation result ($R_i=0\%$) accordingly yields $R_{sa}=0\%$ for the affected subassembly.
This detailed evaluation of each primitive and subtask enables the specific identification of the bottleneck in a selected method.

\textbf{Force evaluation:}
Fig. \ref{fig:baseline_force_metrics} shows the force energy and force smoothness metrics for each primitive.
The variances in the metrics show the influence of uncertainty in real environments on the solution approach.
They are a measure of how well the chosen solution approach can deal with uncertainties in a specific primitive.
Large variances indicate non-robust behavior, while small variances indicate robust behavior.
The \textit{Gearwheel\_Teeth\_2}, \textit{Positioning\_Pin\_5}, \textit{Screw\_M5x16\_3} and \textit{Screw\_M5x16\_4} components demonstrate the highest deviation in force energy in the $z$-direction.
This is partly due to uncertainty in the alignment of objects caused by slipping or unstable grasps.
The baseline does not account for these uncertainties, as it has been deliberately chosen to be straightforward.
Therefore, jamming occurs, resulting in high forces being recorded in the assembly direction and leading to an increased $E_z$ value.
The highest values for $E_{xy}$ are observed for the \textit{Housing\_Middle} and \textit{Module} components, because the rotation of the housing parts causes forces to be exerted in the $x$ and $y$ directions.

The typical \textbf{failure modes} were slipping, jamming, and movement of the object in the hand (unstable grasp).
Fig. \ref{fig:failure_modes} shows example failure cases during execution of the baseline.
The data collected during execution of the baseline will be made available, which include success and failure cases.
It encompasses the force-torque data occurring during assembly, RGB and depth images taken by the wrist-mounted camera, and robot proprioceptive data.
This comprehensive data set can be utilized by other researchers for developing their own applications and evaluating them with the proposed benchmark.
Furthermore, this benchmark will facilitate the identification of corner cases and assembly failures, such as those depicted in Fig. \ref{fig:failure_modes}.

\begin{figure*}[!t]
    \centering
    \begin{subfigure}{0.24\textwidth}
        \centering
        \captionsetup{justification=centering}
        \includegraphics[width=\textwidth]{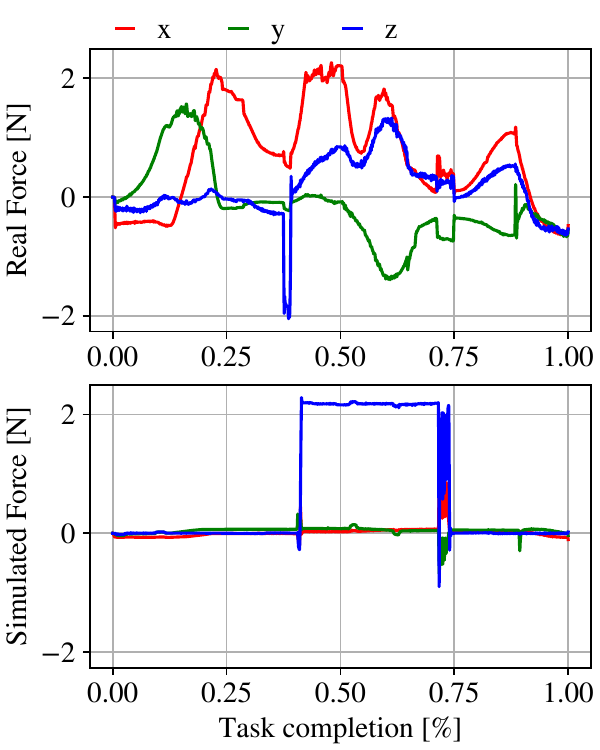}
    \end{subfigure}
    \begin{subfigure}{0.24\textwidth}
        \centering
        \captionsetup{justification=centering}
        \includegraphics[width=\textwidth]{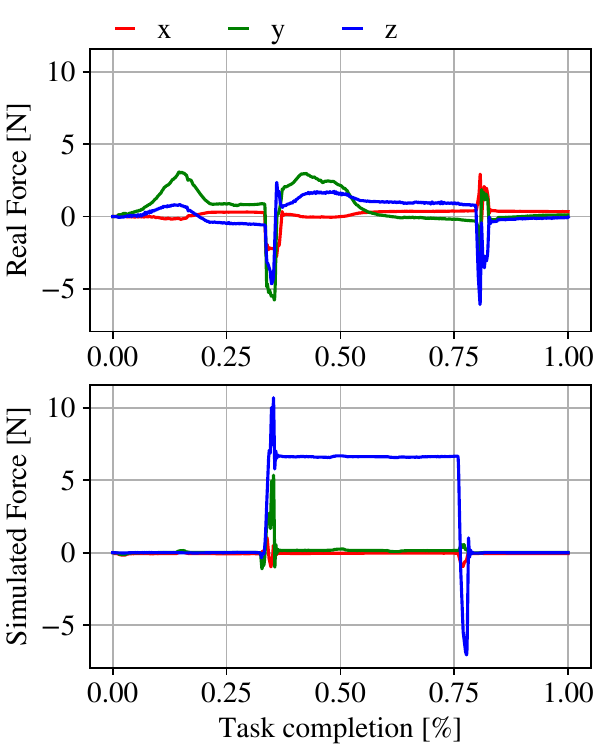}
    \end{subfigure}
    \begin{subfigure}{0.24\textwidth}
        \centering
        \captionsetup{justification=centering}
        \includegraphics[width=\textwidth]{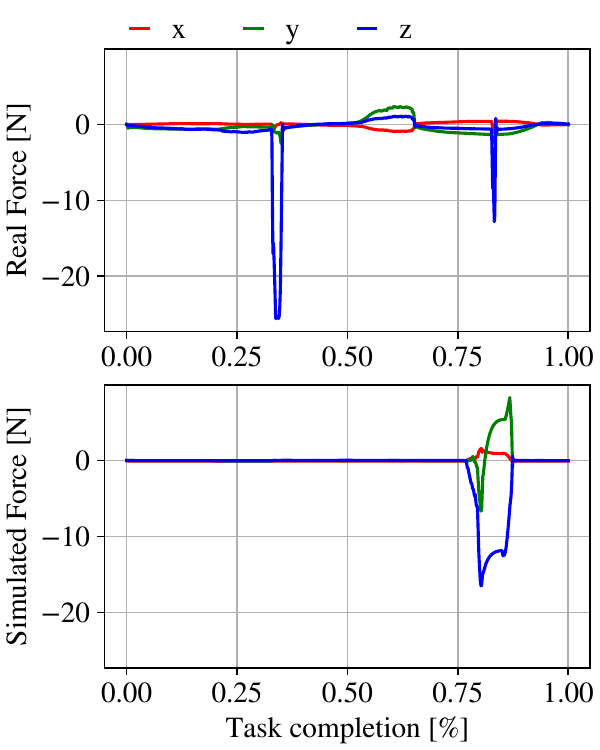}
    \end{subfigure}
    \begin{subfigure}{0.24\textwidth}
        \centering
        \captionsetup{justification=centering}
        \includegraphics[width=\textwidth]{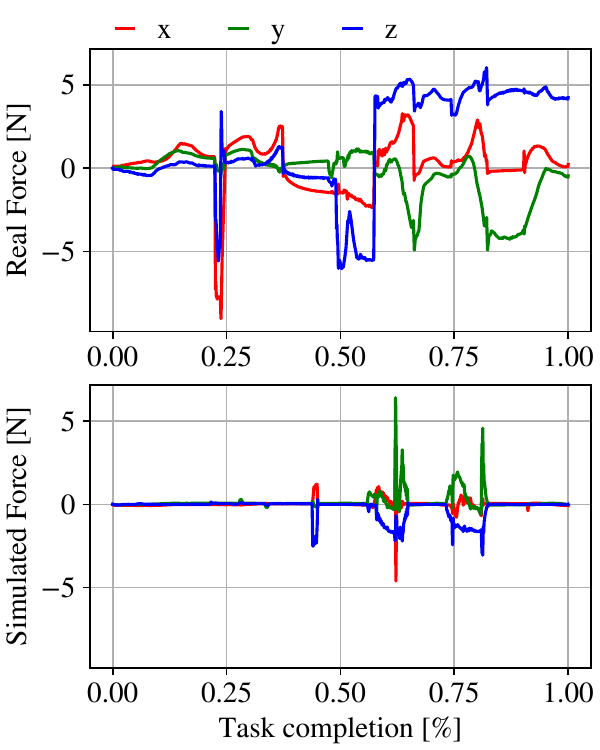}
    \end{subfigure}

    \begin{subfigure}{0.24\textwidth}
        \centering
        \captionsetup{justification=centering}
        \includegraphics[width=\textwidth]{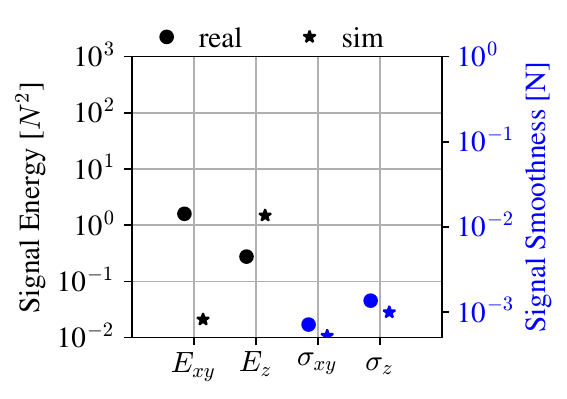}
        \caption{\textit{Gearwheel\_Teeth\_1}}
    \end{subfigure}
    \begin{subfigure}{0.24\textwidth}
        \centering
        \captionsetup{justification=centering}
        \includegraphics[width=\textwidth]{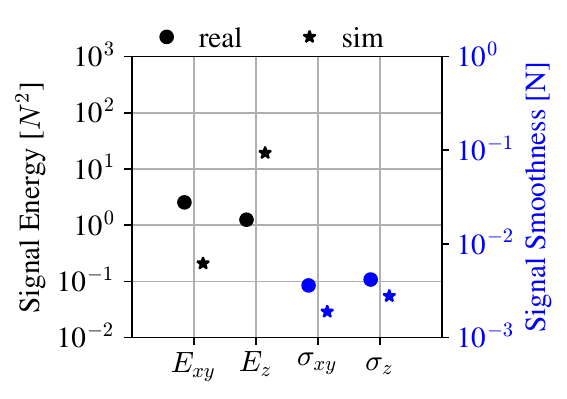}
        \caption{\textit{Housing\_Top}}
    \end{subfigure}
    \begin{subfigure}{0.24\textwidth}
        \centering
        \captionsetup{justification=centering}
        \includegraphics[width=\textwidth]{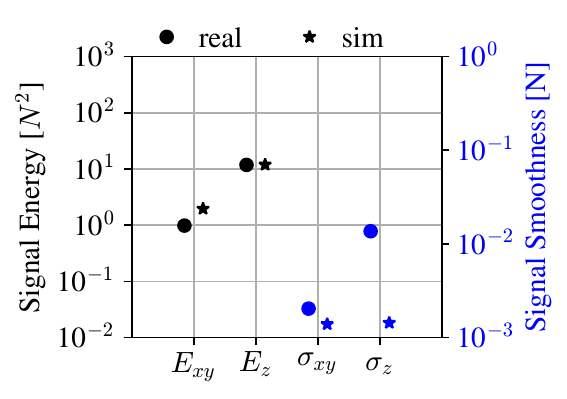}
        \caption{\textit{Plug\_outside\_loose}}
    \end{subfigure}
    \begin{subfigure}{0.24\textwidth}
        \centering
        \captionsetup{justification=centering}
        \includegraphics[width=\textwidth]{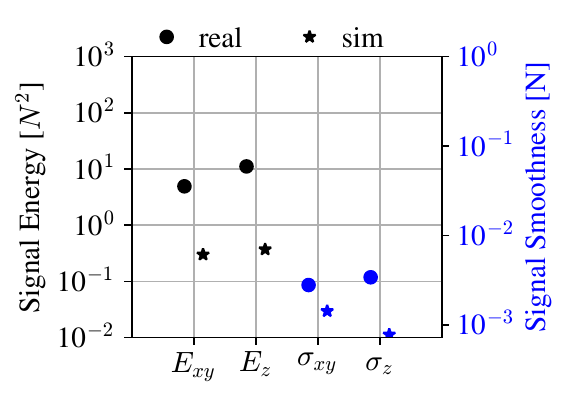}
        \caption{\textit{Screw\_M5x16\_4}}
    \end{subfigure}
    \caption{Comparison of the forces incurred during task execution in the \textbf{real environment} (top row) and the \textbf{simulated environment} (middle row). The bottom row shows the corresponding force interaction metrics.
    (a) insertion of gear wheel 1; (b) alignment and insertion of the top housing onto the middle housing; (c) insertion of the outer plug into the socket on the middle housing; and (d) screwing together of the top and middle housings.}
    \label{fig:sim-real-comparison}
\end{figure*}

\textbf{Sim-Real comparison:}
In practice, carrying out numerous tests to identify the weak points of a selected robot-based assembly approach is very costly.
However, physics-based simulation can achieve this more effectively.
For this to be possible, the simulation must behave in a similar way to reality.
The subsequent section will therefore compare the incurred forces during assembly processes and the resulting interaction metrics between reality and simulation.
Fig. \ref{fig:sim-real-comparison} shows the resulting force profiles after the baseline was executed in the real and simulated environments.
To ensure a fair comparison, the simulation was configured to reproduce the same end-effector velocity profiles as recorded during real execution, so that velocity-dependent force differences are excluded as a source of discrepancy.

Small residual errors in part positioning accumulate between simulation and reality: fixture tolerances, part feeder placement repeatability, and registration offsets mean that the contact point and contact normal at the moment of engagement differ slightly between the two environments.
Even millimeter-level positional deviations change which surface facets make first contact, redistributing force across axes and altering the direction and magnitude of the incurred loads throughout the task.
Also, the in-hand grasp pose of the robot is not perfectly reproduced between simulation and reality.
Slight differences in how the gripper closes on a part, caused by surface texture, finger compliance, and grasp force variability, result in a different relative orientation between the held part and the target.
This orientation error introduces lateral force components that are absent in the idealized simulation grasp, and it alters the moment arm of the contact force, explaining the richer multi-axis force signatures observed in the real profiles.
Additionally, the contact and material parameters in the simulation (stiffness, damping, friction coefficients) are set to nominal values rather than identified from real measurements, causing the simulated contact response to differ in magnitude and duration from the physical one.
Despite these differences, the simulation reproduces the timing and rough magnitude of contact events and remains useful for policy iteration and contact distribution randomization.

\section{Limitations}
The current benchmark has several limitations that future work should address.
First, all assembly operations are restricted to low-force interactions to protect components; high-force operations are not covered.
Second, the baseline evaluation was conducted exclusively at complexity level 1 (part feeder); performance at complexity levels 2 and 3, which require unstructured grasping and planning, remains unreported.
Third, the baseline uses a fixed, manually engineered control pipeline and does not represent flexible or adaptive assembly strategies; the results therefore serve more as a lower bound than as a comprehensive overview of possible approaches.
It is intentionally simple, designed only to demonstrate benchmark functionality and establish a reference point: developing capable methods for complexity levels 2 and 3 is precisely the challenge we invite the community to address using LABIT.
Additionally, the simulation environment is provided as a lightweight MuJoCo setup aligned with the specific LABIT tasks and hardware; other simulators such as robosuite, Genesis, or Isaac Sim are currently not directly supported.
Researchers wishing to use LABIT within those frameworks can do so, as the provided asset files are compatible with those environments.
Finally, due to the use of 3D-printed housing components, dimensional tolerances vary with printer calibration, which may affect reproducibility across different labs.
We provide the 3D-printer settings used on the project website.

\section{Conclusion}
We introduced LABIT, a long-horizon industrial robotic assembly benchmark comprising 59 primitives across seven operation types, a taxonomy-based complexity scoring system, and a paired simulation environment.
LABIT is the first benchmark to span the full assembly sequence of an industrial product, covering hierarchical subassembly dependencies, multi-operation primitives, and three planning complexity levels, enabling reproducible, cross-method comparison on tasks that reflect real manufacturing demands.
The baseline evaluation showed that a classical force-position controller achieves a mean insertion reliability of 82.3\% and a mean screwing reliability of 80\%, while part reorientation remains the critical bottleneck at 0\% for housing-module reorientation.
Jamming and unstable grasps were the dominant failure modes, indicating that future methods should prioritize compliant grasping and in-hand pose estimation.
Future work will aim to reduce the sim-to-real discrepancy through automated contact and material parameter estimation, and to use LABIT as a platform for developing and evaluating adaptive, perception-driven assembly pipelines.
All components, designs, and protocols are publicly available to support community adoption.

\section*{Declaration of generative AI and AI-assisted technologies in the
writing process}
During the preparation of this work the author(s) used Anthropic Claude in
order to enhance writing and sentence accuracy. After using this tool/service, the
author(s) reviewed and edited the content as needed and take(s) full
responsibility for the content of the publication.

\section*{CRediT authorship contribution statement}
\textbf{Constantin Schempp:} Conceptualization, Methodology, Software, Validation, Investigation, Data Curation, Visualization, Writing - Original Draft.
\textbf{Christian Friedrich:} Conceptualization, Writing - Review \& Editing, Supervision, Project administration, Funding acquisition.
\textbf{Luca Kofler:} Methodology, Investigation.
\textbf{Laura Suhm:} Methodology, Investigation.

\section*{Declaration of competing interests}
The authors declare that they have no known competing financial interests or personal relationships that could
have appeared to influence the work reported in this paper.

\section*{Data availability}
Data is available on the project website:\\
{\color{blue}{\href{https://irp-hka.github.io/robotic-assembly-benchmark/}{https://irp-hka.github.io/robotic-assembly-benchmark/}}}.

\section*{Acknowledgement}
This work was supported by the German Federal Ministry of Research, Technology and Space (BMFTR), Grant Reference Number 03FHP105.

\bibliographystyle{elsarticle-num}
{\let\newline\relax
\def\urlprefix#1#2{}
\bibliography{mybibfile}}

\end{document}